\documentclass[10pt,twocolumn,letterpaper]{article}

\usepackage[pagenumbers]{cvpr} % To force page numbers, e.g. for an arXiv version

\usepackage{graphicx}
\usepackage{amsmath}
\usepackage{amssymb}
\usepackage{booktabs}
\usepackage{tabularray}
\usepackage{geometry}
\usepackage{multirow}
\usepackage{array}
\usepackage{ragged2e}
\usepackage{threeparttable}
\usepackage{pifont}
\usepackage{tikz}
\usepackage{placeins}

\usepackage[pagebackref,breaklinks,colorlinks]{hyperref}

\usepackage[capitalize]{cleveref}
\crefname{section}{Sec.}{Secs.}
\Crefname{section}{Section}{Sections}
\Crefname{table}{Table}{Tables}
\crefname{table}{Tab.}{Tabs.}

\def\cvprPaperID{*****} % *** Enter the CVPR Paper ID here
\def\confName{CVPR}
\def\confYear{2022}

\begin{document}

\title{M3D: Advancing 3D Medical Image Analysis with Multi-Modal Large Language Models}  

\author{Fan Bai$^{1,2}$, Yuxin Du$^{1}$, Tiejun Huang$^{1,3}$, Max Q.-H. Meng$^{2,4 \dag}$, Bo Zhao$^{1 \dag}$\\ 
\small $^{1}$Beijing Academy of Artificial Intelligence
\small $^{2}$The Chinese University of Hong Kong  \\
\small $^{3}$Peking University \quad 
\small $^{4}$Southern University of Science and Technology\\
\small $^{\dag}$Corresponding author: Bo Zhao $<$bozhaonanjing@gmail.com$>$, Max Q.-H. Meng $<$max.meng@ieee.org$>$
}

\maketitle

%%%%%%%%% ABSTRACT
\begin{abstract}
Medical image analysis is essential to clinical diagnosis and treatment, which is increasingly supported by multi-modal large language models (MLLMs). However, previous research has primarily focused on 2D medical images, leaving 3D images under-explored, despite their richer spatial information. This paper aims to advance 3D medical image analysis with MLLMs. To this end, we present a large-scale 3D multi-modal medical dataset, \textbf{M3D-Data}, comprising 120K image-text pairs and 662K instruction-response pairs specifically tailored for various 3D medical tasks, such as image-text retrieval, report generation, visual question answering, positioning, and segmentation. Additionally, we propose \textbf{M3D-LaMed}, a versatile multi-modal  large  language model for 3D medical image analysis. Furthermore, we introduce a new 3D multi-modal medical benchmark,  \textbf{M3D-Bench}, which facilitates automatic evaluation across eight tasks. Through comprehensive evaluation, our method proves to be a robust model for 3D medical image analysis, outperforming existing solutions. All code, data, and models are publicly available at: \url{https://github.com/BAAI-DCAI/M3D}.
\end{abstract}

%%%%%%%%% BODY TEXT
\section{Introduction}
\label{sec:intro}
Medical scenarios \cite{pei2023mmdata} encompass a vast amount of multi-modal information, including patient information, diagnostic reports, and medical images of various modalities. Diagnostic reports paired with medical images provide accurate and detailed descriptions, findings, and diagnoses, considered as high-quality annotations. These medical images and texts are saved into the database along with the physician's diagnostic workflow at scale and without additional cost. How to make full use of these image and text data to build a medical image diagnosis model is a key issue.

In recent work \cite{li2023blip2,liu2023llava, zhu2023minigpt, openai2023gpt4}, multi-modal large language models (MLLMs) have demonstrated remarkable performance in various multi-modal tasks, effectively integrating image and text data. By combining the perceptual abilities of vision models \cite{radford2021clip, EVA-CLIP, zhai2023siglip} with the generative capabilities of large language models (LLMs) \cite{touvron2023llama, zheng2023vicuna, thoppilan2022lamda, du2022glm, chowdhery2022palm,chatgpt}, MLLMs have garnered significant attention from researchers, particularly in medical image analysis. Existing medical MLLMs \cite{zhang2023pmcvqa, li2023llavamed, wu2023generalist, zhang2024biomedgpt} fine-tune publicly available 2D MLLMs on medical image and text data to achieve tasks such as image-text retrieval, report generation, and visual question answering. These models are proposed as powerful tools for understanding and reasoning about 2D medical images. However, when confronted with widespread 3D medical images, such as CT and MRI, which contain rich spatial information, these methods often struggle, either requiring costly slice-by-slice analysis or failing outright.

In this work, we focus on 3D medical images and extend the use of MLLMs to analyze them. To this end, we collect a large-scale 3D multi-modal medical dataset, M3D-Data, comprising 120K image-text pairs and 662K instruction-response pairs covering various diseases and tasks. This dataset is the largest public 3D multi-modal medical dataset to date and can advance related research. Additionally, we propose M3D-LaMed, a versatile 3D MLLM for medical image analysis. It can perform tasks such as image-text retrieval, report generation, and visual question answering, and also includes tasks such as vision language positioning and segmentation for the first time. Leveraging a pre-trained 3D vision encoder under a CLIP-like\cite{radford2021clip} strategy and an efficient 3D spatial pooling perceiver, it can understand and reason about 3D images directly. For the first time, M3D-LaMed is combined with a 3D promptable segmentation model to achieve referring expression segmentation of 3D medical images. Furthermore, to evaluate the model's ability in 3D medical analysis, we propose a multi-modal medical benchmark, M3D-Bench, which includes 8 tasks covering various aspects of 3D medical image analysis. This is the first comprehensive benchmark in 3D medical image analysis. In addition to traditional metrics, we have introduced LLM-based evaluation, allowing M3D-Bench to automatically and accurately evaluate the model's performance.

In summary, our contributions are as follows:
\begin{itemize}
    \item Establish M3D-Data, a large-scale 3D medical dataset containing 120K image-text pairs and 662K instruction-response pairs.
    
    \item Propose M3D-LaMed, a versatile MLLM for 3D medical image analysis, applied to various 3D multi-modal tasks.
    
    \item Create M3D-Bench, a comprehensive 3D multi-modal benchmark for 8 tasks.
\end{itemize}

\begin{table*}[tb]
\centering
\caption{Comparisons of M3D-Data with other datasets. M3D-Data contains image-text pairs (M3D-Cap) and instruction-response pairs (M3D-VQA, M3D-RefSeg, and M3D-Seg), the largest 3D medical dataset, and involves the most tasks. VQA: Visual Question Answering under closed- and open-ended. ITR: Image-Text Retrieval. RG: Report Generation. REC: Referring Expression Comprehension. REG: Referring Expression Generation, SS: Semantic Segmentation. RES: Referring Expression Segmentation.}
\label{tab1}
\setlength{\tabcolsep}{5mm}{
\begin{threeparttable} 
\begin{tabular}{@{}lccrr@{}}
\toprule
Datasets            & Types & Tasks             & Images  & Texts     \\ \midrule
VQA-Med\cite{ben2019vqa}            & 2D    & VQA               & 3,200   & 12,792    \\
MIMIC-CXR\cite{johnson2019mimic} & 2D    & ITR, RG           & 377,110 & 227,835   \\
PMC-OA\cite{lin2023pmc} & 2D    & ITR, RG           & -       & 1,646,592 \\
PMC-VQA\cite{zhang2023pmcvqa} & 2D    & VQA               & 149,075 & 226,946   \\ \midrule
RP3D-Caption\cite{wu2023radfm} & 3D    & ITR, RG            & 51K       & -       \\
RP3D-VQA\cite{wu2023radfm} & 3D    & VQA               & -       & 142K      \\
\textbf{M3D-Cap}    & 3D    & ITR, RG           & 120,092 & 42,496    \\
\textbf{M3D-VQA}    & 3D    & VQA               & 96,170  & 509,755   \\
\textbf{M3D-RefSeg} & 3D    & REC, REG, SS, RES & 210     & 2,778     \\
\textbf{M3D-Seg}    & 3D    & REC, REG, SS, RES & 5,772   & 149,196*         \\ \bottomrule
\end{tabular}
\begin{tablenotes}
\item[*] In segmentation datasets, the number of texts can be linked to semantic masks.
\end{tablenotes}
\end{threeparttable} 
}
\end{table*}

\section{Related Work}
\subsection{Medical Multi-Modal Datasets}
In medical scenarios \cite{pei2023mmdata}, rich images of various modalities and texts are available. However, previous works \cite{ben2019vqa,johnson2019mimic} have difficulty constructing large-scale medical multi-modal datasets due to privacy and restrictions. Inspired by CLIP\cite{radford2021clip}, PMC-OA\cite{lin2023pmc} obtained image and text data from medical papers through web crawling, resulting in 1.6M 2D image-text pairs. Additionally, MedMD\cite{wu2023radfm} aims to achieve multiple objectives: building 2D and 3D medical models, integrating public 2D medical datasets, and crawling 3D image and text data from medical professional websites. One of its 3D datasets, RP3D\cite{wu2023radfm}, comprises 51K 3D image-text pairs and 142K VQA data generated from LLMs. In our work, we primarily focus on constructing large-scale 3D medical datasets by crawling medical professional websites. M3D-Data includes 120K 3D image-text pairs and 662K instruction-response pairs generated through an automatic and low-cost data generation pipeline. Furthermore, M3D-Data's M3D-Seg component collects nearly 6K 3D images from 25 public medical segmentation datasets, facilitating tasks such as vision language positioning and segmentation. In summary, M3D-Data is the largest 3D medical multi-modal dataset, supporting various tasks, as shown in Table~\ref{tab1}.

\subsection{Medical MLLMs}
Medical MLLMs \cite{li2023llavamed, wu2023generalist, zhang2024biomedgpt} are typically fine-tuned from powerful 2D open-source MLLMs using medical multi-modal datasets. For instance, LLaVA-Med \cite{li2024llavamed}, Med-PaLM M\cite{tu2024medpalmm}, and Med-Flamingo\cite{moor2023medflamingo} are based on models such as LLaVA\cite{liu2023llava}, PaLM-E\cite{driess2023palme}, and Flamingo\cite{alayrac2022flamingo}, respectively. The availability of large-scale datasets like PMC-VQA \cite{zhang2023pmcvqa} has enabled training medical MLLMs from scratch, although initially limited to 2D images. While RadFM \cite{wu2023radfm} supports both 2D and 3D images, it is primarily used for text generation tasks such as VQA and has poor performance. In our work, M3D-LaMed serves as a generalist MLLM for 3D medical image analysis. It handles not only text generation tasks like report generation and VQA but also pioneers vision tasks, like vision language positioning and segmentation in 3D medical images, which are crucial for identification and localization in medical image analysis.

\section{Dataset}
The goal of M3D-Data is to provide a data source for various 3D multi-modal medical tasks. M3D-Data comprises 120K image-text and 662K instruction-response pairs covering 8 tasks, as shown in Table~\ref{tab1}.

\begin{figure*}[tb]
\centering
\includegraphics[width=0.9\textwidth]{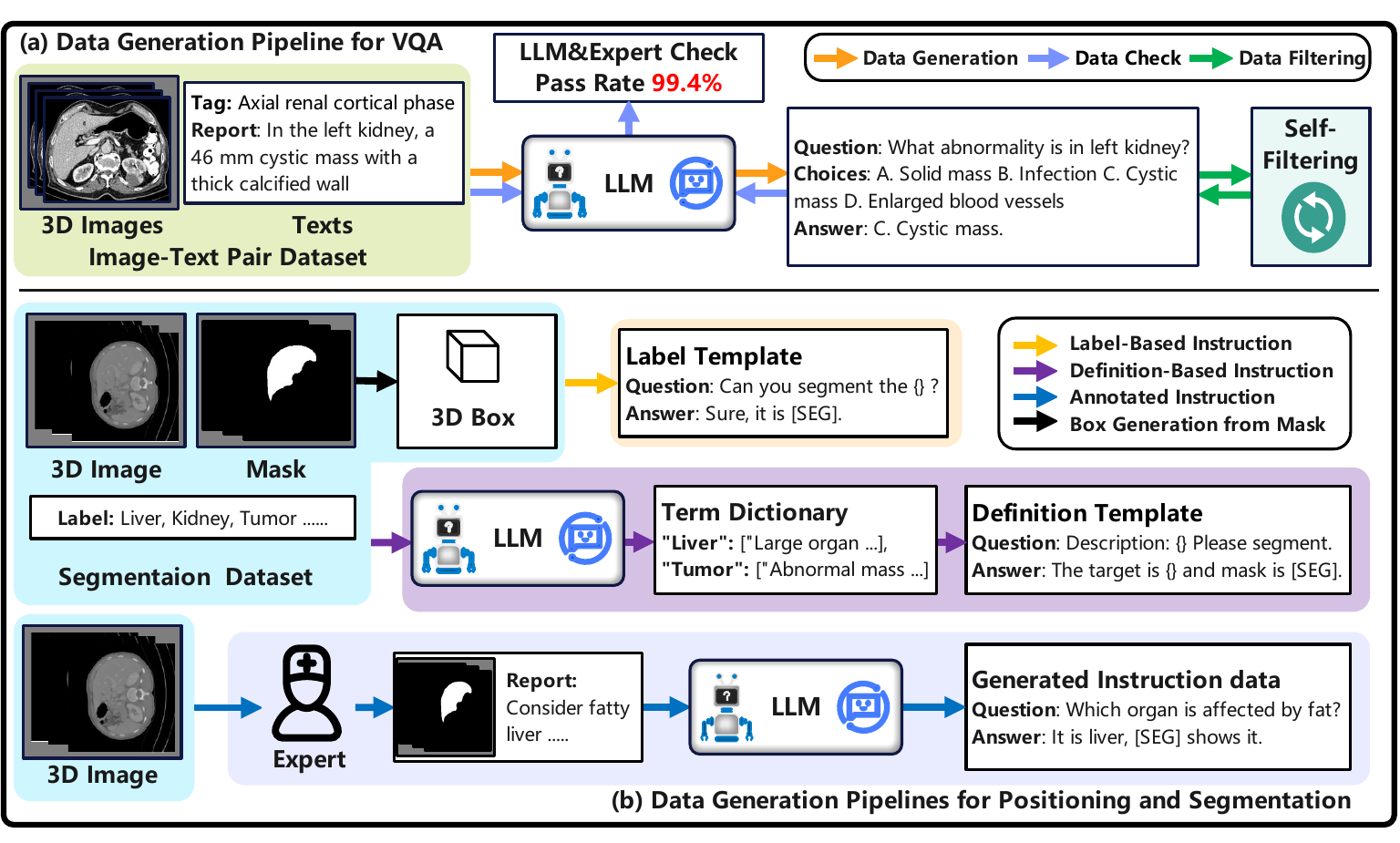}
\caption{The generation pipelines for M3D-Data. (a) In the VQA data generation pipeline, we employ LLM to generate five types of questions from medical reports using a prompt-based method. Subsequently, we eliminate dirty data through self-filtering and check the test set by LLM and experts, achieving a pass rate of 99.4\%. (b) In positioning and segmentation data generation pipelines, three feasible methods were implemented to construct image-mask-text triplets, including label-based instruction, definition-based instruction, 
and annotated instruction. The box coordinates required for positioning tasks can be generated directly from masks.} 
\label{fig1}
\end{figure*}

\subsection{Image-Text Pair Data}
Healthcare institutions, such as hospitals, maintain extensive repositories of medical images and diagnostic reports, often accompanied by text descriptions. However, disclosing such comprehensive image-text datasets presents challenges due to the sensitive nature of patient data and privacy concerns. To avoid these privacy concerns, we collect medical images and reports from publicly accessible professional medical websites \footnote{Radiopaedia: \url{https://radiopaedia.org/}\label{rad}}. Specifically, each patient case in our dataset includes multiple images along with their corresponding reports, which are supplemented by meticulously peer-reviewed captions provided by experts from Radiopaedia\textsuperscript{\ref{rad}}. Focusing on 3D CT data due to its critical role in medical image analysis, particularly in diagnosing, positioning, and measuring lesions throughout the body, we successfully compiled a large-scale 3D medical image-text pair dataset, M3D-Cap, comprising 120K pairs. M3D-Cap supports tasks such as image-text retrieval and report generation.

\subsection{Instruction-Response Pair Data}
Instruction-response pairs data contains pairs of instructions or questions and corresponding responses or answers. These data are commonly used in various multi-modal understanding and generation tasks, such as VQA, vision language positioning, and segmentation. The sum of the instruction data for these tasks is 662K. 

\begin{figure*}[tb]
\centering
\includegraphics[width=0.75\textwidth]{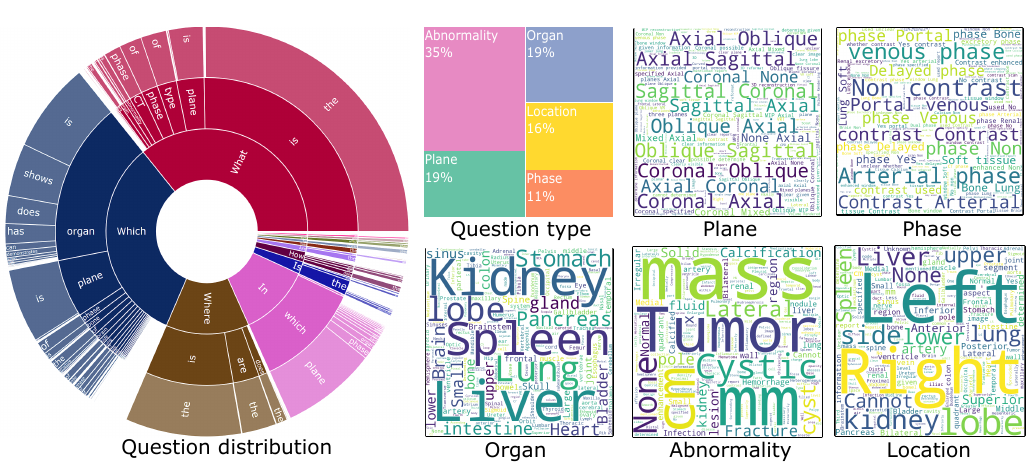}
\caption{The data statistics of M3D-VQA on five question types. What, which, and where are 3 typical questions. Samples of 5 topics are displayed in word clouds.} 
\label{fig2}
\end{figure*}

\subsubsection{VQA Data}
The acquisition and annotation of VQA data in medical scenarios often involve substantial expenses, primarily due to the requirement for specialized medical expertise. To mitigate these costs, we leverage public LLMs to analyze existing diagnostic reports and generate instruction-response pairs using a prompt-based approach, as illustrated in Figure \ref{fig1}(a). Additionally, we apply self-filtering techniques to eliminate noise data based on predefined rules. Following this process, the test set undergoes validation by both LLM-based analysis and expert review, resulting in an impressive pass rate of 99.4\% (13729/13814). Considering the cost implications associated with utilizing ChatGPT \cite{chatgpt}, we opt for the free and robust Qwen-72B \cite{qwen} for data generation. Specifically, we generate multiple-choice questions covering 5 key topics extracted from each diagnostic report: plane, phase, organ, abnormality, and location. The data statistics of M3D-VQA, including topic samples, are illustrated in Figure \ref{fig2}. Adopting the multiple-choice format offers convenience for both open- and closed-ended evaluation, facilitating comprehensive assessment.

\subsubsection{Positioning and Segmentation Data}
Vision language positioning and segmentation require the integration of images, texts, and referring regions, typically represented as boxes and masks corresponding to the tasks. To streamline data handling, we adopt a unified data format consisting of image-mask-text triplets. This format provides resources for positioning tasks and segmentation by converting masks into coordinates of 3D boxes. 
However, the rarity of lesion mask annotations during diagnosis in healthcare institutions makes the creation of a 3D image-mask-text pair dataset more costly due to the necessity for detailed region annotations. Therefore, we developed datasets using three distinct approaches, as depicted in Figure \ref{fig1}(b): \textbf{(1) Label-base instruction data:} Directly created from image-mask pairs of public segmentation datasets using label templates. \textbf{(2) Definition-based instruction data:} Constructed using a term dictionary containing multiple related definitions and descriptions generated by LLMs, with subsequent utilization of definition templates for instruction data creation. \textbf{(3) Annotated instruction:} Expert annotation of text descriptions about referring regions to produce image-mask-text triplets. Additionally, we leverage the low-cost yet powerful LLM, Qwen-72B, to enhance textual content and generate comprehensive instruction data based on annotations. In approaches (1) and (2), we compile the joint segmentation dataset, M3D-Seg, by amalgamating densely annotated 3D CT segmentation datasets obtained from public repositories, as shown in detail in Appendix Table \ref{stab2}. Meanwhile, in approach (3), we annotate a subset, M3D-RefSeg, from the Totalsegmentator\cite{wasserthal2023totalsegmentator} dataset.

\begin{figure*}[tb]
\centering
\includegraphics[width=0.9\textwidth]{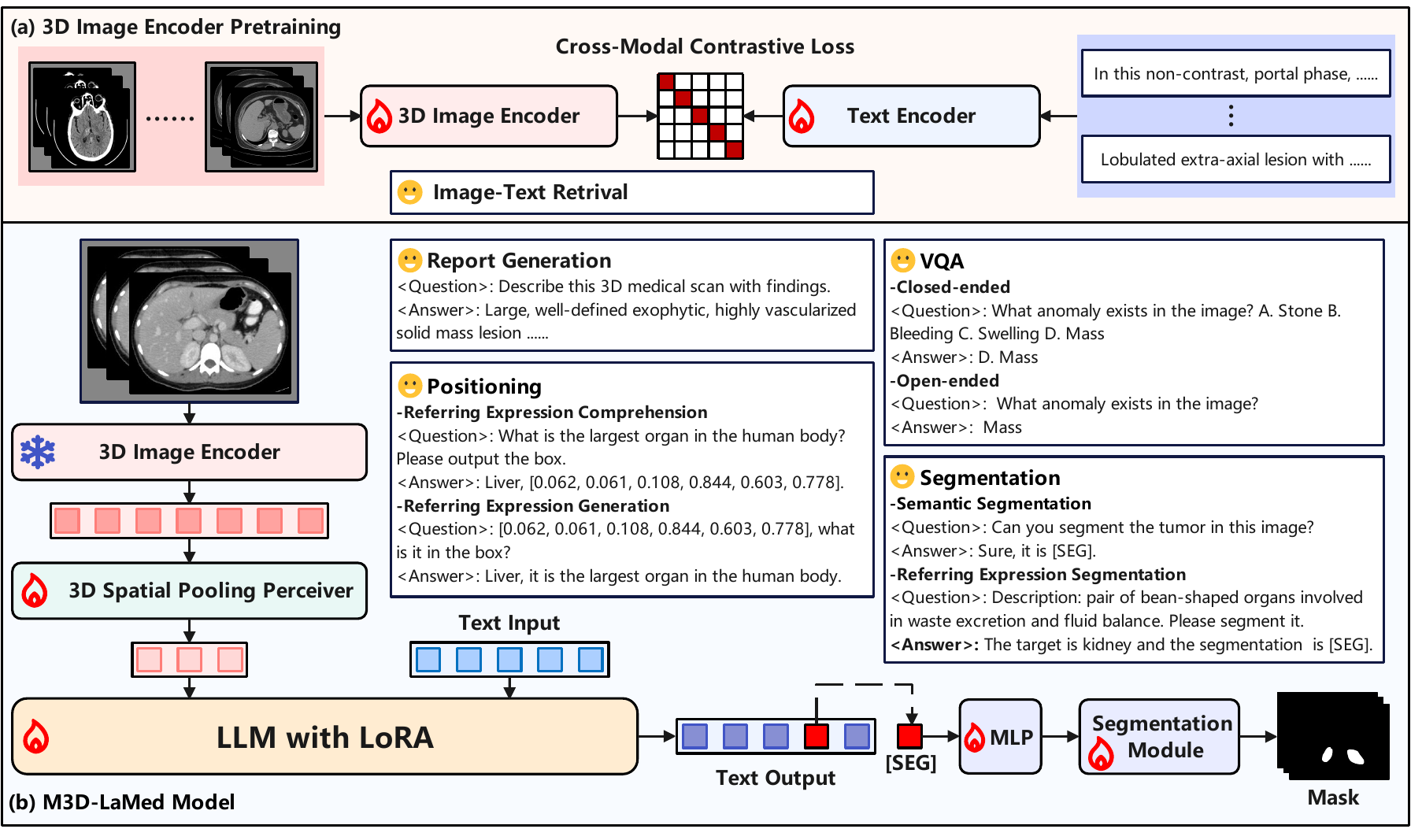}
\caption{Overview of our M3D-LaMed model. (a) The 3D image encoder is pre-trained by cross-modal contrastive learning loss with image-text pairs, performing image-text retrieval. (b) In the M3D-LaMed model, 3D medical images are fed into a pre-trained 3D image encoder and an effective 3D spatial pooling perceiver to produce refined embeddings inserted into LLM. The output [SEG] token is used as a prompt to employ the promptable 3D medical segmentation model, thereby obtaining the 3D mask output. With our M3D-Data, M3D-LaMed can be leveraged on various 3D medical tasks.} 
\label{fig3}
\end{figure*}

\section{Method}
Due to unreliable 3D encoders, we train the vision encoder from scratch. As depicted in Figure \ref{fig3} (a), we pre-train the 3D medical vision encoder using a CLIP-like\cite{radford2021clip} strategy on M3D-Cap. Subsequently, we introduce end-to-end tuning to integrate 3D information into the LLM using instruction data, ensuring seamless interaction between the vision and language, shown in Figure \ref{fig3} (b).

\subsection{Model Architecture}
\subsubsection{3D Image Encoder}
Given a specific 3D image $I \in R^{C \times D \times H \times W}$, $C, D, H, W$ represent the channel, depth, height, and width respectively, we derive the image embedding $v = E_{img}(I) \in R^{n \times d}$. Here, $E_{img}$ denotes the image encoder, $n$ represents the image token numbers, and $d$ denotes the token dimensions. For versatility and generality, we utilize the 3D Vision Transformer (3D ViT) \cite{dosovitskiy2021vit} as the vision encoder. The 3D ViT comprises an $N$-layer transformer with attention mechanisms. Each layer operates on patches extracted from the input image, where the patch size is $P_D*P_H*P_W$. We can import standard 3D ViT directly from the MONAI \footnote{MONAI: \url{https://monai.io/}} library.
    
\subsubsection{3D Perceiver}
Since 3D images inherently possess high dimensions and numerous tokens, direct input into an LLM results in significant computational costs. To alleviate this challenge, we propose a straightforward and efficient 3D spatial pooling perceiver designed to reduce the number and dimensions of embeddings, shown in Appendix Figure \ref{sfig1}. Firstly, the output tokens from the vision encoder are reconstructed into 3D space for pooling. This step effectively decreases token numbers while retaining the original spatial information. Secondly, we employ a series of Multi-Layer Perceptrons (MLPs) to adjust the embedding dimensions, aligning them with the dimensions required by the LLM. By implementing these steps, the 3D perceiver not only mitigates computational costs but also ensures the preservation of spatial information.

\subsubsection{LLM}
Large language models trained on extensive natural language corpora offer generalized embedding representations and powerful generation capabilities. In our study, we utilize the LLaMA-2-7B model directly as our base LLM due to its proven effectiveness in capturing linguistic patterns and generating coherent text across various domains.

\subsubsection{Promptable Segmentation Module}
Inspired by LISA\cite{lai2023lisa}, we leverage the capabilities of MLLMs to implement referring expression segmentation using a promptable segmentation module. Specifically, if an [SEG] token is present in the output tokens, we extract the last layer embedding of the [SEG] token as a feature. Subsequently, we map this feature into a prompt to drive the segmentation module through an MLP, ultimately producing the segmentation mask. For our implementation, we opt for SegVol\cite{du2023segvol} as the promptable segmentation module due to its robust performance and compatibility with our framework.

\begin{table*}[tb]
\centering
\caption{Comparison on image-text retrieval. Our model outperforms previous models at various difficulty levels. IR means image-to-text retrieval. TR means text-to-image retrieval. R@1, R@5, and R@10 denote recall at 1,5,10.}
\label{tab2}
\setlength{\tabcolsep}{2mm}{
\begin{tabular}{@{}cl|cccc|cccc@{}}
\toprule
\multicolumn{2}{c|}{Methods}                    & \multicolumn{4}{c|}{PMC-CLIP\cite{lin2023pmc}} & \multicolumn{4}{c}{Our}       \\ \midrule
\multicolumn{2}{c|}{Test samples}               & 100      & 500      & 1000    & 2000    & 100   & 500   & 1000  & 2000  \\ \midrule
\multicolumn{1}{c|}{\multirow{3}{*}{IR}} & R@1  & 9.00     & 4.40     & 1.90    & 1.15    & 64.00 & 39.60 & 27.30 & 19.10 \\
\multicolumn{1}{c|}{}                    & R@5  & 28.00    & 12.80    & 7.60    & 4.35    & 95.00 & 76.20 & 61.10 & 47.45 \\
\multicolumn{1}{c|}{}                    & R@10 & 45.00    & 18.80    & 12.10   & 7.60    & 99.00 & 87.20 & 76.10 & 62.25 \\ \midrule
\multicolumn{1}{c|}{\multirow{3}{*}{TR}} & R@1  & 18.00    & 7.60     & 4.60    & 3.15    & 70.00 & 40.40 & 26.60 & 18.45 \\
\multicolumn{1}{c|}{}                    & R@5  & 47.00    & 20.20    & 13.00   & 8.55    & 95.00 & 74.20 & 61.80 & 47.30 \\
\multicolumn{1}{c|}{}                    & R@10 & 59.00    & 31.00    & 19.80   & 13.55   & 98.00 & 87.00 & 75.30 & 62.15 \\ \bottomrule
\end{tabular}
}
\end{table*}

\begin{table*}[tb]
\centering
\caption{Comparison on report generation. In the LLM-based metric, we use Qwen-72B to score generations and references on an absolute value of 0-100 based on the content overlap.}
\setlength{\tabcolsep}{2mm}{
\label{tab3}
\begin{tabular}{@{}lccccc@{}}
\toprule
Method       & BLEU  & ROUGE & METEOR & BERT-Score & LLM-Based \\ \midrule
RadFM\cite{wu2023radfm} & 12.23 & 16.49 & 11.57  & 87.93      & 4.32 \\
Our (Linear) & 14.49 & 19.25 & 14.11  & 88.32      & 7.12      \\
Our (MLP)    & 15.15 & 19.55 & 14.38  & 88.46     & 8.49     \\ 
\bottomrule
\end{tabular}
}
\end{table*}

\subsection{Vison Encoder Pre-training}
Due to the lack of robust 3D medical image encoders, we adopt the architecture and training approach of CLIP\cite{radford2021clip} to pre-train on M3D-Cap. As shown in Figure \ref{fig3} (a), we utilize cross-modal contrastive learning loss in pre-training. The vision encoder is pre-trained from scratch, while the text encoder utilizes a pre-trained BERT \cite{devlin2019bert} as initialization.

\subsection{MLLM Training}
Upon obtaining the pre-trained 3D medical vision encoder, we integrate it into the LLM using a 3D perceiver for end-to-end training. Our training process consists of two main steps. Initially, we freeze the vision encoder and the LLM, focusing on fine-tuning only the 3D perceiver using image-text pairs. Subsequently, we fine-tune the vision encoder, 3D perceiver, LLM, and segmentation module using instruction data. If the [SEG] token is present in the output tokens, segmentation training is performed using Dice and BCE loss similar to \cite{du2023segvol}. Considering the training cost and retaining the original knowledge of LLM, we adopt the LoRA \cite{hu2021lora} strategy for parameter-efficient fine-tuning on the LLM.

\section{Evaluation and Benchmark}
M3D-Bench is a comprehensive and accurate evaluation benchmark that covers 8 tasks from 5 major tasks: image-text retrieval, report generation, VQA, positioning, and segmentation.

\subsection{Evaluation on Image-Text Retrieval}
In 3D image-text retrieval, the model aims to match images and texts from a dataset based on their similarity, typically involving two tasks: text-to-image retrieval (TR) and image-to-text retrieval (IR). For evaluation, we selected a high-quality subset of 2000 pairs from M3D-Cap as the test set. This set is further categorized into four difficulty levels: easy (100 pairs), medium (500 pairs), difficult (1000 pairs), and very difficult (2000 pairs), based on the size of the retrieving data pool.
Evaluation metrics include recall at ranks 1, 5, and 10 for IR and TR, assessing the model's ability to retrieve relevant images or texts from top-ranked results.

\begin{table*}[tb]
\caption{Comparison on 3D closed-ended VQA. Our model outperforms other 3D MLLM by a large margin in five types of questions.}
\label{tab4}
\centering
\setlength{\tabcolsep}{2mm}{
\begin{tabular}{@{}lcccccc@{}}
\toprule
Methods      & Plane & Phase & Organ & Abnormality & Location & Mean  \\ \midrule
RadFM\cite{wu2023radfm} & 19.65 & 28.70 & 16.80  & 18.92   & 14.88 &19.79  \\
Our& 98.80 & 79.75 & 74.75 & 66.65 & 58.94 &75.78 \\
\bottomrule
\end{tabular}
}
\end{table*}

\begin{table*}[tb]
\caption{Evaluation on 3D open-ended VQA. Our model outperforms the other MLLM in five types of problems and four metric evaluations.}
\label{tab5}
\centering
\begin{tabular}{@{}llcccccc@{}}
\toprule
Method                        & Metric     & Plane & Phase & Organ & Abnormality & Location & Mean  \\ \midrule
\multirow{4}{*}{RadFM\cite{wu2023radfm}} & BLEU       & 14.24 & 14.25 & 14.24 & 15.64       & 23.58    & 16.39 \\
                              & ROUGE      & 25.40 & 25.41 & 25.38 & 25.38       & 29.09    & 26.13 \\
                              & METEOR     & 20.62 & 20.63 & 20.61 & 20.60       & 24.19    & 21.33 \\
                              & BERT-Score & 92.68 & 92.04 & 86.79 & 85.84 & 86.26    & 88.72 \\ \midrule
\multirow{4}{*}{Our}    & BLEU       & 98.37 & 74.41 & 34.20 & 15.91       & 24.00    & 49.38 \\
                              & ROUGE      & 98.42 & 78.63 & 37.87 & 19.27       & 27.74    & 52.39 \\
                              & METEOR     & 49.20 & 63.58 & 23.78 & 12.83       & 18.50    & 33.58 \\
                              & BERT-Score & 99.47 & 95.55 & 88.97 & 86.08       & 87.60    & 91.53 \\ \bottomrule

\end{tabular}
\end{table*}

\subsection{Evaluation on Report Generation}
In report generation, the model produces text reports based on information extracted from 3D medical images. For evaluation, we use the same test set from the image-text retrieval task. We set 2000 high-quality image-text pairs as the basic test set and set 1000 image-text pairs as a small test set to facilitate user evaluation. All experiments are tested on the basic test set. 
Due to the challenge of assessing content accuracy between generated reports and human references, we employ two types of metrics: traditional and LLM-based metrics. Traditional metrics include BLEU (Bilingual Evaluation Understudy)\cite{papineni2002bleu}, ROUGE (Recall-Oriented Understudy for Gisting Evaluation)\cite{lin2004rouge}, METEOR (Metric for Evaluation of Translation with Explicit Ordering)\cite{banerjee2005meteor}, and BERT-Score\cite{zhang2019bertscore}. These metrics measure aspects like n-gram overlap or variations to quantify text similarity, albeit with limited semantic understanding.
LLM-based metrics utilize LLMs with robust semantic understanding as evaluators, such as the public Qwen-72B. These metrics assess the aspects mentioned in generation and human reference and calculate the percentage of correct or partially matched aspects, assigning a score from 0 to 100.

\begin{figure*}[t]
\centering
\includegraphics[width=0.5\textwidth]{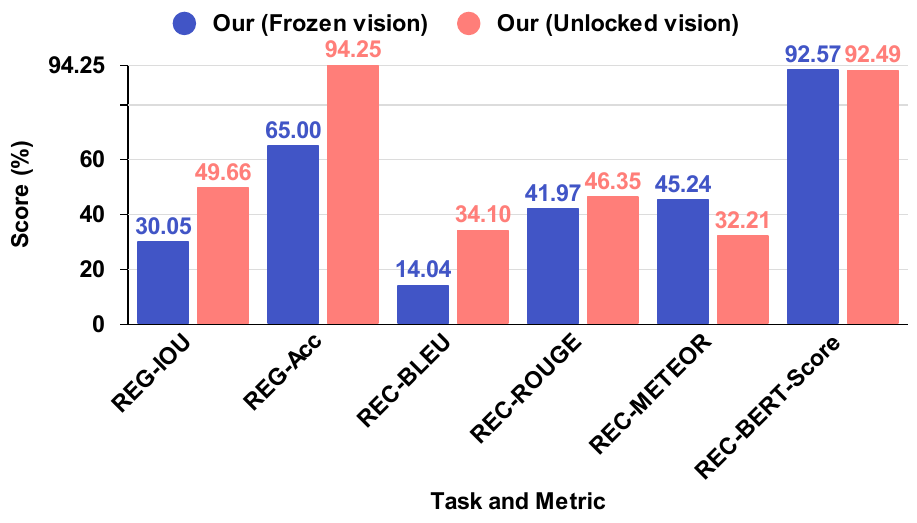}
\caption{Comparison on 3D positioning (REG \& REC) with frozen or unlocked vision encoder in fine-tuning. Unlocking the vision encoder can improve the performance of the output box, the REG task.} 
\label{fig4}
\end{figure*}

\begin{table*}
\centering
 \caption{Comparison on 3D segmentation. Our model not only surpasses previous methods in semantic segmentation tasks but also achieves RES tasks previously unattainable. ACT-1K: AbdomenCT-1K. TS: ToalSegmentator. CTOrg: CT-Oran.}
  \label{tab6}
    \begin{tabular}{@{}lcccccc@{}}
      \toprule
      \multirow{2}{*}{Methods} & \multicolumn{3}{c}{Semantic Segmentation} & \multicolumn{3}{c}{Referring Expression Segmentation} \\ \cmidrule(l){2-7} 
      & ACT-1K\cite{Ma-2021-AbdomenCT-1K}         & TS\cite{wasserthal2023totalsegmentator}         & CTOrg\cite{rister2020ct}        & ACT-1K \cite{Ma-2021-AbdomenCT-1K}  & TS \cite{wasserthal2023totalsegmentator}   & CTOrg \cite{rister2020ct}   \\ \midrule
      SegVol \cite{du2023segvol}                  & 79.06        & 44.28        & 77.78       & -     & -      & -      \\
      Our (Linear)             & 73.64        & 58.30        & 81.27       & 73.63 & 58.50  & 83.48  \\
      Our (MLP)                & 73.37        & 59.70        & 81.01       & 73.36 & 72.38 & 83.49 \\ \bottomrule
    \end{tabular}
\end{table*}

\begin{table}
\centering
\caption{Ablation study on closed-ended VQA. }
\label{tab7}
\begin{tabular}{@{}ccccc@{}}
\toprule
\begin{tabular}[c]{@{}c@{}}Vision\\ pre-train\end{tabular} & \begin{tabular}[c]{@{}c@{}}Spatial\\ pooling\end{tabular} & MLP                        & \begin{tabular}[c]{@{}c@{}}Unlocked\\ vision\end{tabular} & \begin{tabular}[c]{@{}c@{}}VQA\\ Mean\end{tabular} \\ \midrule
\ding{56}   & \ding{52}   & \ding{52} &   \ding{56}  & 71.13 \\
\ding{52} & \ding{56} & \ding{52}& \ding{56} & 72.87         \\
\ding{52} & \ding{52}  & \ding{56}&   \ding{56} & 73.50      \\
\ding{52} & \ding{52} & \ding{52}& \ding{56} & 74.96 \\
\ding{52} & \ding{52} & \ding{52}& \ding{52} & 75.78  
\\ \bottomrule
\end{tabular}
\end{table}

\subsection{Evaluation on VQA}
The VQA tasks involve generating text answers related to given images and questions, which typically fall into two categories: open-ended and closed-ended. In open-ended VQA, the model generates answers without constraints, while closed-ended VQA restricts the acceptable answers to a predefined set of choices.
To evaluate both forms, we organize M3D-VQA into multiple-choice questions with four answer choices (A, B, C, D). Subsequently, we provide two test sets with different sizes: the basic test set and the small test set. The basic test set comprises 2000 3D medical images and 13,791 question-answer pairs covering five types. The small test set comprises 1000 3D medical images and 5000 question-answer pairs covering five types. The results of this paper are based on the basic test set, and the small test set is prepared for faster and low-cost evaluation by users. After self-filtering, which removes erroneous or low-quality data, a combination of LLM and expert check is employed to identify issues, achieving a pass rate of 99.4\%.
For closed-ended VQA, questions and choices serve as prompt inputs, and answers are supervision signals. The accuracy is directly computed by answers and choices. For open-ended VQA, the model is trained using questions and corresponding correct answers. Evaluation involves measuring the similarity between model-generated texts and references using metrics such as BLEU, ROUGE, METEOR, and BERT-Score.

% \begin{table}[tb]
% \centering
% \caption{Evaluation on 3D vision-language positioning}
% \label{tab6}
% \begin{tabular}{@{}ccccccc@{}}
% \toprule
% \multirow{2}{*}{Method} & \multicolumn{2}{c}{REG} & \multicolumn{4}{c}{REC}             \\
%                         & IOU        & Acc        & BLEU  & ROUGE & METEOR & BERT-Score \\ \midrule
% Our                     & 30.05      & 65.00      & 14.04 & 41.97 & 45.24  & 92.57      \\ \bottomrule
% \end{tabular}
% \end{table}

\subsection{Evaluation on Positioning}
Positioning is crucial in vision language tasks \cite{chen2023shikra}, especially those involving input and output boxes. Tasks involving output boxes, like Referring Expression Comprehension (REC) \cite{kazemzadeh2014referitgame, mao2016generation}, aim to localize a target object in an image based on a referring expression. In contrast, tasks with input boxes, such as Referring Expression Generation (REG) \cite{liu2017referring}, require the model to generate a description of a specific region given an image and a location box. In our M3D-RefSeg and M3D-Seg, the masks are converted into box coordinates representing the maximum bounding rectangle $(x_1, y_1, z_1, x_2, y_2, z_2)$. In evaluation, 20\% of the data in AbdomenCT-1K \cite{Ma-2021-AbdomenCT-1K} from M3D-Seg is used as the test set. For tasks involving output boxes, we computed the Intersection over Union (IOU) between the output boxes and ground truth as the localization metric. The IOU score greater than 0.2 is considered as hitting the target, and positioning accuracy is reported accordingly. For tasks with input boxes, BLEU, ROUGE-L, METEOR, and BERT-Score are calculated as metrics for evaluating generated descriptions and answers.

\begin{figure*}[tb]
\centering
\includegraphics[width=0.9\textwidth]{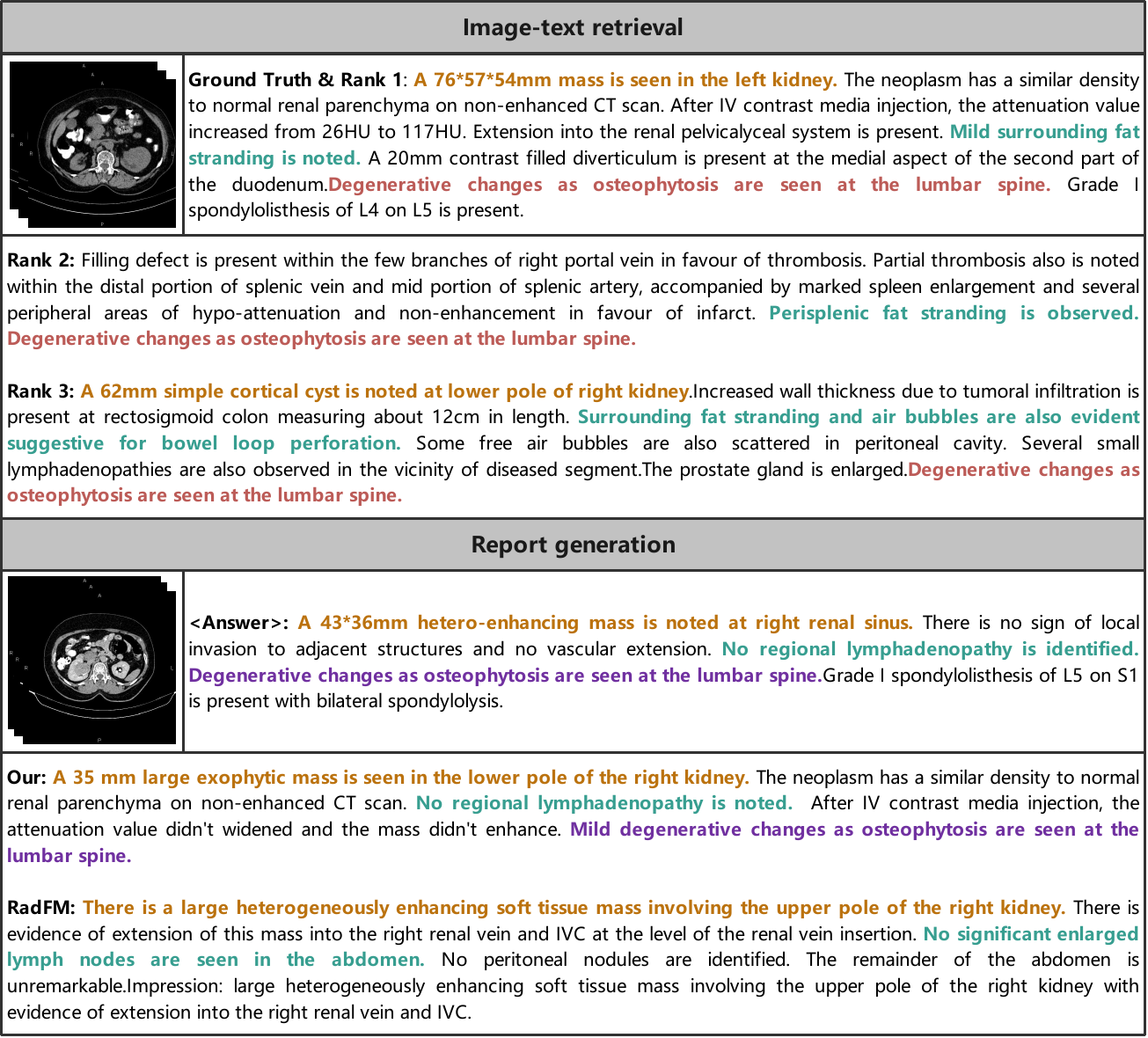}
\caption{Qualitative comparisons with our model and ground truth on image-text retrieval and report generation. The same colors marked in prediction and answer represent similar content.} 
\label{fig5}
\end{figure*}

\begin{figure*}[tb]
\centering
\includegraphics[width=0.9\textwidth]{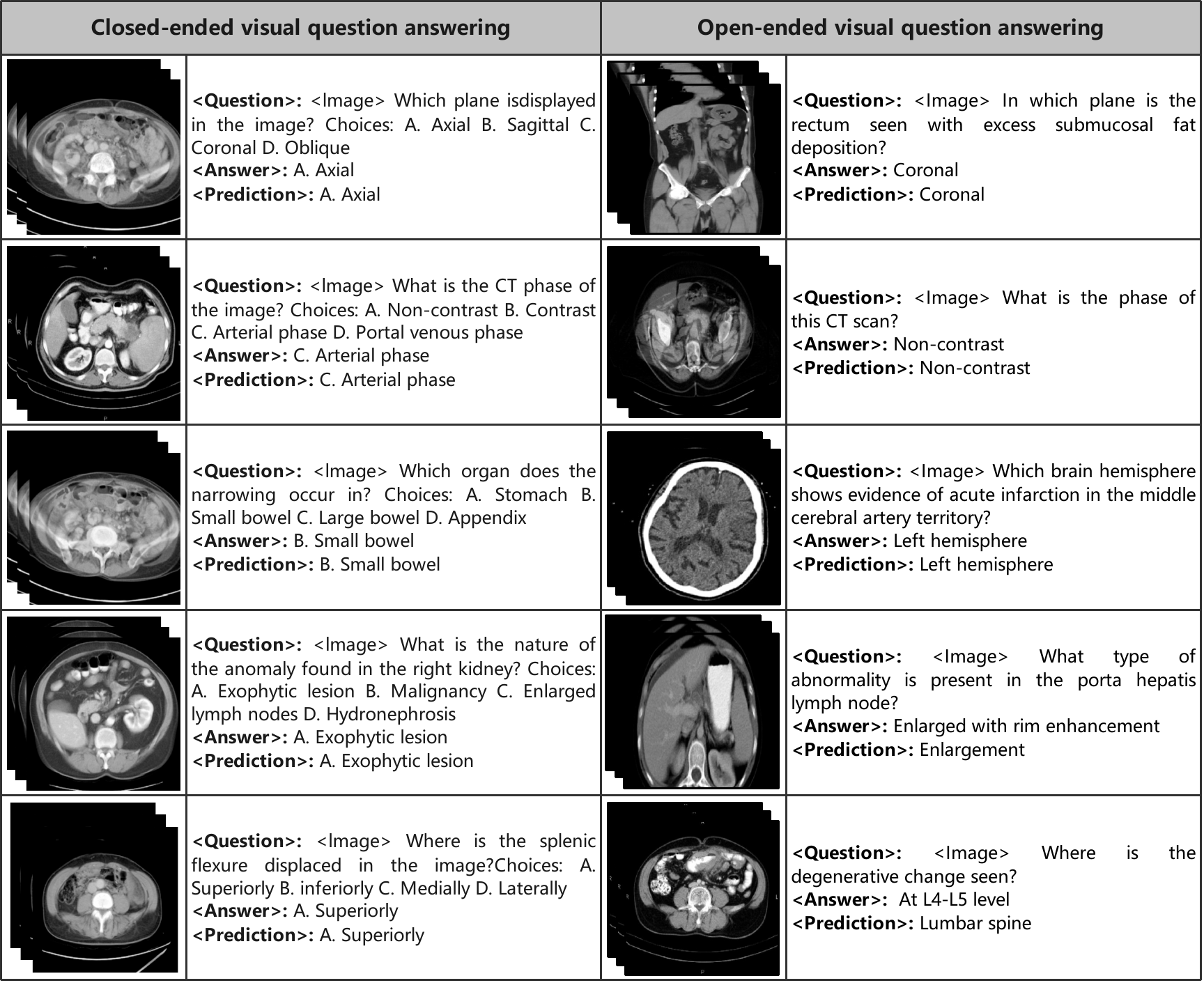}
\caption{Qualitative comparisons on closed-ended and open-ended VQA. Closed-ended VQA requires the model to pick the correct answer from the input options, while open-ended VQA has no option as input.} 
\label{fig6}
\end{figure*}

\begin{figure*}[tb]
\centering
\includegraphics[width=1\textwidth]{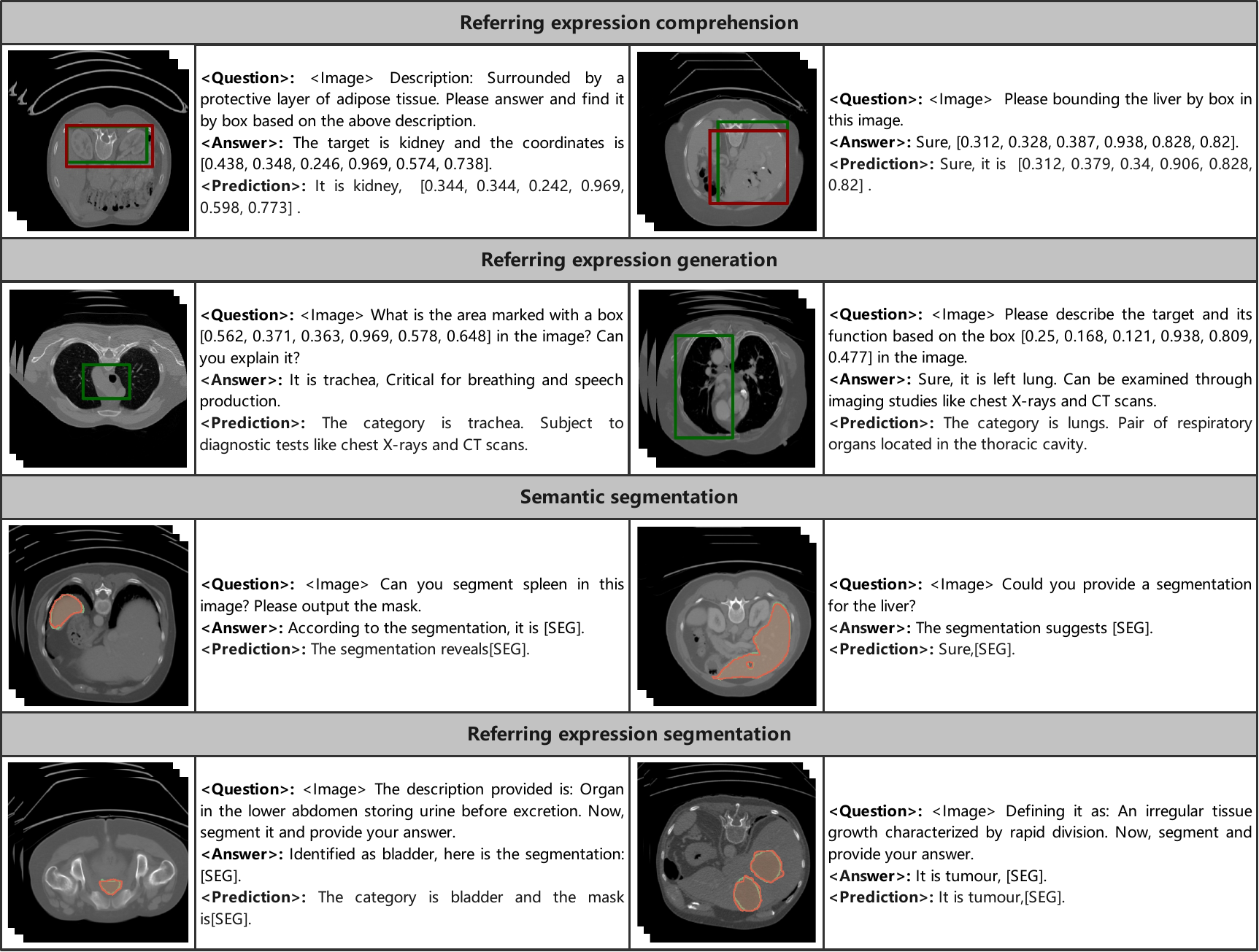}
\caption{Qualitative comparisons on positioning and segmentation. The green box (mask) in the 3D image indicates the ground truth, while the red box (mask) indicates the prediction.} 
\label{fig7}
\end{figure*}

\begin{figure*}[t]
\centering
\includegraphics[width=0.95\textwidth]{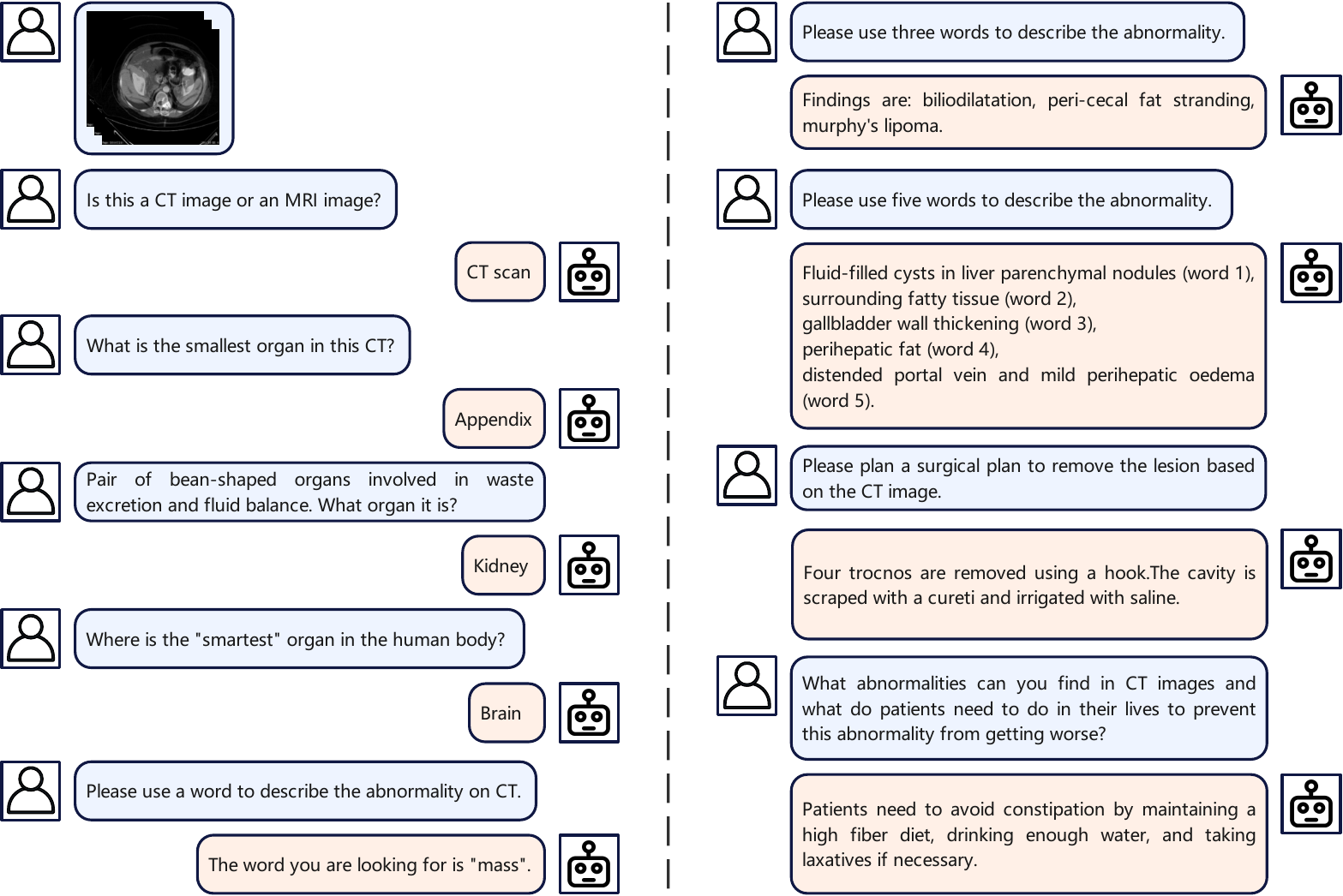}
\caption{Case study on out-of-distribution (OOD) questions. We test the M3D-LaMed model on OOD dialogue, which means that all questions are \textbf{NOT} related to our training data. We found that M3D-LaMed has strong generalization and can produce reasonable answers to OOD questions instead of gibberish. In each set of conversations, the avatar and questions on the left come from the user, and the avatar and answers on the right come from M3D-LaMed.} 
\label{fig8}
\end{figure*}

\subsection{Evaluation on Segmentation}
The segmentation task is vital in 3D medical image analysis due to its recognition and localization capabilities. To address various text prompts, segmentation is categorized into semantic segmentation and referring expression segmentation. For semantic segmentation, the model generates segmentation masks based on semantic labels. Referring expression segmentation involves segmenting targets described by natural language expressions using the model. In M3D-Seg, 20\% of the data from AbdomenCT-1K\cite{Ma-2021-AbdomenCT-1K}, Totalsegmnetator \cite{wasserthal2023totalsegmentator}, and CT-Organ \cite{rister2020ct} is allocated as the test set for both semantic segmentation and referring expression segmentation. Traditionally, Dice is employed as a metric.

\section{Experiments}
\subsection{Implementation Details}
\label{implementation_details}
We apply Min-Max Normalization consistently to preprocess the 3D CT images as input. Additionally, we resize and crop the 3D images to a standardized size of $32\times256\times256$. Our 3D vision encoder adopts a 3D ViT, using 12 layers of transformers with a patch size of $4\times16\times16$. The output embeddings are $2048\times768$, representing 2048 tokens with 768 feature dimensions. After our 3D spatial pooling perceiver, the final vision tokens fed to the LLM are $256\times768$. We utilize the LLaMA-2-7B as an LLM base and load pre-trained parameters.

In vision encoder pre-training, we employ BERT with 12 layers of transformers as the text encoder, with a maximum text length of 128. The $[CLS]$ tokens of both the vision encoder and text encoder serve as the global feature representation, and a linear layer is utilized for cross-modality contrastive training projection. Additionally, we adopt a batch size of $6\times8$ for parallel training across 8 GPUs, with a learning rate of $10^{-4}$ and warm-up and cosine decay schedules.

There are two stages in MLLM training. Initially, we freeze the vision encoder and LLM and solely fine-tune the 3D perceiver with image-text pairs, employing a batch size of $12\times8$, a learning rate of $10^{-4}$, and applying warm-up and cosine decay. Specifically, we explore two situations: 1-layer linear and 2-layer MLP in the perceiver. Subsequently, we fine-tune the vision encoder, 3D perceiver, LLM, and segmentation modules with instruction data, utilizing a batch size of $12\times8$, a learning rate of $2\times10^{-5}$, and employing warm-up and cosine decay. The segmentation module uses the parameters of SegVol\cite{du2023segvol} as initialization. We consistently apply the parameter-efficient LoRA method for LLM fine-tuning, with LoRA parameters set to $r=16$, $\alpha=32$, and a dropout rate of 0.1. The maximum context length is defined as 512.

All our models are trained using the AdamW \cite{kingma2014adam, loshchilov2017decoupled} optimizer and leverage the bf16 Mixed-precision Training Strategy enabled by DeepSpeed to optimize the training process. Implementation is carried out in PyTorch, and training occurs in parallel across 8 NVIDIA A100 GPUs with 80 GB memory.

\subsection{Results on Image-Text Retrieval}
Since 3D CLIP-like models do not exist in medical image analysis, we consider using 2D models as baselines. During the evaluation, we sample 10 equally spaced 2D slices for each 3D image along the depth dimensions. When using a 2D model to perform retrieval on slices, we identify the 3D image whose slice has the highest similarity to the target. We attempted to use CLIP\cite{radford2021clip}, but it performed poorly in the medical domain with scores close to 0. Hence, we chose PMC-CLIP\cite{lin2023pmc} as the baseline model in the medical domain.
In Table \ref{tab2}, we find that the retrieval performance of our model far exceeds that of the 2D PMC-CLIP model at various difficulty levels, which lacks spatial information. For example, under the simplest settings (100 test samples and R@10 metric), our model outperforms PMC-CLIP by 54\% in IR. Even under the most difficult setting (2000 samples and R@1 metric), our model exceeds PMC-CLIP by 17.95\% in IR. Figure \ref{fig5} qualitatively demonstrates that text retrieved based on images has similar content.

\subsection{Results on Report Generation}
Table \ref{tab3} compares performance with RadFM and our model using Linear or MLP in the perceiver. Regardless of whether using traditional or LLM-based metrics, our model outperforms RadFM in report generation. For example, our model with MLP exceeds RadFM by 2.92\% in terms of BLEU score. Based on the more accurate evaluation using the LLM Qwen-72B, our model even surpasses RadFM by 4.17\%. Moreover, our model's performance with MLP is slightly better than with Linear. Figure \ref{fig5} qualitatively demonstrates the powerful generative capabilities of our model, in which the reports we generate have more in common with the correct answers.

\subsection{Results on VQA}
We assess the performance of the M3D-LaMed model on closed-ended and open-ended VQA utilizing the M3D-VQA dataset. In Table \ref{tab4}, for closed-ended VQA, our model performs far better than RadFM on all five types of problems. Referring to Table \ref{tab5} for open-ended VQA, our model also significantly outperforms RadFM. In Figure \ref{fig6}, we qualitatively demonstrate our model's understanding capabilities on open-ended VQA and closed-ended VQA.

\subsection{Results on Positioning}
Figure \ref{fig4} depicts the evaluation of the 3D vision-language positioning task, considering two subtasks, REG (Referring Expression Generation) for output with a box and REC (Referring Expression Comprehension) for input with a box, respectively. We compare the frozen and unlocked vision encoders during fine-tuning. We found that the unlocked vision encoder significantly improved the REG task. For example, The accuracy increased by 29.25\%, which requires generating box output from the 3D image. However, in the REC task, the unlocked vision encoder does not achieve consistent performance improvements. In Figure \ref{fig7}, we provide a qualitative showcase of our model's prowess in the vision language positioning task.

\subsection{Results on Segmentation}
In Table \ref{tab6}, we evaluate the 3D segmentation task, considering SS (Semantic Segmentation) and RES (Referring Expression Segmentation). Leveraging the comprehension capabilities of multi-modal large models, our performance surpasses SegVol in several aspects. Furthermore, we possess RES capabilities that SegVol lacks. In Figure \ref{fig7}, we showcase the segmentation capabilities of our model qualitatively.

\subsection{Ablation Study}
In Table \ref{tab7}, we conduct ablation studies in our closed-set VQA task on four aspects: vision pre-train, spatial pooling, MLP, and unlocked vision. Omitting vision pre-training involves training from scratch. Omitting spatial pooling entails direct pooling on sequence tokens. Omitting MLP entails replacing it with a single linear layer. Omitting unlocked vision involves freezing the vision encoder during fine-tuning. Through detailed ablation experiments, we found that every part of the model is irreplaceable, and using visual pre-training as a starting point and unlocking the visual encoder in fine-tuning is a better training solution.

\subsection{Case Study on OOD Questions}
We aim to investigate the generalization ability of our model, specifically its capacity to answer OOD questions beyond the scope of the training set. To this end, we devise unconventional problems, illustrated in Figure \ref{fig8}. For instance, in a CT scan of the chest and abdomen, our model identifies the appendix as the smallest organ, a concept absent from the training data. Similarly, when presented with the grammatically awkward query "smartest organ", the model aptly responded with "Brain", despite this phrase not being part of the training data.
While our dataset already contains questions describing anomalies, we imposed stricter constraints, such as limiting queries to one, three, and five words. Remarkably, our model successfully addresses these constraints, even though it had not been explicitly trained for such scenarios. Furthermore, when presented with queries related to surgical planning or seeking life advice, the model generates pertinent responses, demonstrating its adaptability beyond the confines of the training data. In summary, our M3D-LaMed model exhibits robust generalization capabilities for OOD problems. This proficiency stems from our approach of performing lightweight LoRA fine-tuning on LLM rather than full-parameter fine-tuning, which preserves the LLM's original understanding and knowledge. By leveraging LLM's inherent capabilities and fine-tuning on new multi-modal datasets, MLLM emerges with enhanced professional and generalization capabilities. Hence, developing a medical MLLM based on a robust LLM foundation proves indispensable.

\section{Conclusion}
In conclusion, our study advances 3D medical image analysis with MLLM. Specifically, we construct a large-scale 3D multi-modal medical dataset, M3D-Data, comprising 120K 3D image-text pairs and 662K instruction-response pairs tailored for 3D medical tasks. Additionally, we propose M3D-LaMed, a generalist model handling image-text retrieval, report generation, visual question answering, positioning, and segmentation. Furthermore, we introduce a comprehensive benchmark, M3D-Bench, meticulously designed for eight tasks. We assert that our approach establishes a robust foundation for MLLMs to comprehend the vision and language of 3D medical scenarios. The availability of our data and code will facilitate further exploration and application for 3D medical MLLM in future research.

%%%%%%%%% REFERENCES
{\small
\bibliographystyle{ieee_fullname}
\bibliography{egbib}
}

\clearpage
\appendix
\setcounter{figure}{0} % Reset figure counter
\setcounter{table}{0} % Reset table counter

Some supplementary materials are provided additionally to validate the proposed approach. The supplementary materials include:
\begin{itemize}
    \item[\textbullet] Implementation details of our models and datasets, which are not included in our manuscript, in Section \ref{sec1};
    \item[\textbullet] Detailed prompts and templates for data generation, data check, model evaluation, and task instruction, in Section \ref{sec2};
    \item[\textbullet] More qualitative analysis about 8 tasks, in Section \ref{sec3}.
\end{itemize}

\section{Implementation Details}
\label{sec1}

\subsection{Modules}

\begin{figure}[htb]
\centering
\includegraphics[width=0.35\textwidth]{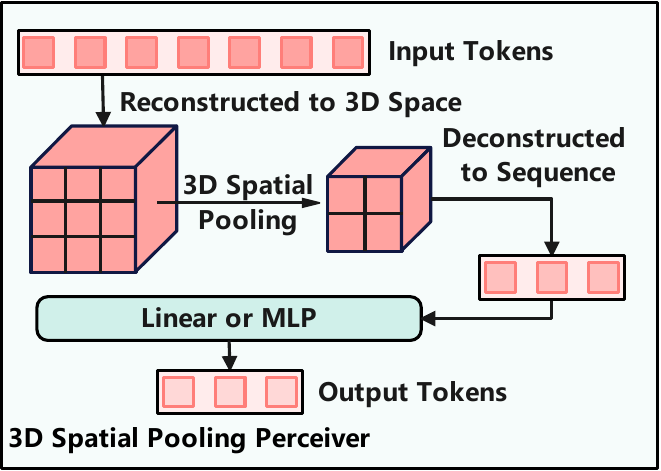}
\caption{The architecture of the 3D spatial pooling perceiver. (1) To reduce the number of tokens, input tokens are first reconstructed into 3D space for pooling and then deconstructed into sequence tokens. (2) To align the dimensions, A projection layer, like linear or MLP, converts the sequence tokens to the same dimension as the LLM. The effect of different projection layers, linear and MLP, on performance is reported in our experiment.} 
\label{sfig1}
\end{figure}

We introduce our 3D spatial pooling perceiver in Figure \ref{sfig1}. This architecture simplifies and efficiently reduces the number of 3D image tokens while aligning dimensions, enabling the injection of 3D image information into LLM.

Detailed module parameters of the M3D-LaMed model are presented in Table \ref{tab1}. Specifically, we explore two situations: 1-layer linear and 2-layer MLP. The overall model parameters amount to 6.9 billion, considerably smaller than RadFM\cite{wu2023radfm}, which has 13 billion parameters. Although the LLM base constitutes 97\% of all parameters, fine-tuning LLM with just LoRA during training is exceptionally cost-effective.

\begin{table}[htb]
\centering
\caption{The parameters of each module in our M3D-LaMed. Linear means using a 1-layer linear in the perceiver for dimensional alignment, while MLP means using a 2-layer MLP for alignment. We utilize 3D ViT with a 12-layer transformer as a 3D image encoder, LLaMA-2-7B\cite{touvron2023llama} as an LLM base, and SegVol\cite{du2023segvol} as a segmentation module.} 
\label{stab1}
\setlength{\tabcolsep}{2mm}{
\begin{tabular}{@{}lr@{}}
\toprule
Modules                                                                             & Parameters \\ \midrule
3D Image Encoder                                                                    & 87.4M      \\
\begin{tabular}[c]{@{}l@{}}3D Spatial Pooling Perceiver\\ (Linear/MLP)\end{tabular} & 3.1M/19.9M \\
LLM with LoRA                                                                       & 6.7B       \\
Segmentation Module                                                                 & 117.3M     \\ \midrule
All                                                                                 & 6.9B       \\ \bottomrule
\end{tabular}
}
\end{table}

\begin{table*}[htb]
\centering
\caption{Detailed dataset composition in M3D-Seg. M3D-Seg contains 5,772 labeled 3D CTs from 25 public datasets. All data, download links, and processing scripts will be made public. CAT: Category.} 
\label{stab2}
\setlength{\tabcolsep}{1mm}{
\begin{tabular}{@{}llrrrr@{}}
\toprule
Datasets                                                                                     & Anatomical Targets                             & CAT & Train          & Test           & All           \\ \midrule
3D-IRCADB\cite{soler20103d}                                                 & Liver and liver tumor                          & 47       & 16             & 4              & 20            \\
FLARE22\cite{FLARE22}                   & Thoracic and abdominal organs                  & 13       & 40             & 10             & 50            \\
AbdomenCT-1k\cite{Ma-2021-AbdomenCT-1K}                                     & Liver, kidney, spleen, pancreas            & 4        & 800            & 200            & 1000         \\
AMOS22\cite{ji2022amos}                                                     & Abdominal organs                               & 15       & 192            & 48             & 240           \\
BTCV\cite{landman2015miccai}                                                & Abdominal organs                               & 13       & 24             & 6              & 30            \\
CHAOS\cite{CHAOS2021, CHAOSdata2019, kavur2019}                             & Abdominal organs                               & 1        & 16             & 4              & 20            \\
CT-ORG\cite{rister2019ct, rister2018ct, bilic2023liver, clark2013cancer}    & Organs of the body & 6        & 112            & 28             & 140           \\
HaN-Seg\cite{podobnik2023han}                                               & Organs of the head and neck                    & 30       & 33             & 9              & 42            \\
KiPA22\cite{he2021meta, he2020dense, shao2011laparoscopic, shao2012precise} & Kidney, renal tumor, artery, vein          & 4        & 56             & 14             & 70            \\
KiTS19\cite{heller2020state}                                                & Kidney and kidney tumor                        & 2        & 168            & 42             & 210           \\
KiTS23\cite{heller2023kits21}                                               & Kidney, kidney tumor and cyst          & 3        & 391            & 98             & 489           \\
LUNA16\cite{setio2017validation}                                            & Left lung, right lung, trachea             & 3        & 710            & 178            & 888           \\
MSD-Colon\cite{simpson2019large}                                            & Colon tumor                                    & 1        & 100            & 26             & 126           \\
MSD-HepaticVessel\cite{simpson2019large}                                    & Hepatic vessel and liver tumor                 & 2        & 242            & 61             & 303           \\
MSD-Liver\cite{simpson2019large}                                            & Liver and liver tumor                          & 2        & 104            & 27             & 131           \\
MSD-Lung\cite{simpson2019large}                                             & Lung tumor                                     & 1        & 50             & 13             & 63            \\
MSD-Pancreas\cite{simpson2019large}                                         & Pancreas and pancreas tumor                    & 2        & 224            & 57             & 281           \\
MSD-Spleen\cite{simpson2019large}                                           & Spleen                                         & 1        & 32             & 9              & 41            \\
Pancreas-CT\cite{roth2016data, roth2015deeporgan, clark2013cancer}          & Pancreas                                       & 1        & 65             & 17             & 82            \\
QUBIQ\cite{QUBIQ_2021}                                                     & Kidney, pancreas and lesion          & 3        & 65             & 17             & 82            \\
SLIVER07\cite{heimann2009comparison}                                        & Liver                                          & 1        & 16             & 4              & 20            \\
TotalSegmentator\cite{wasserthal2022totalsegmentator}                       & Organs of the whole body                       & 104      & 962            & 241            & 1203          \\
VerSe19\cite{sekuboyina2021verse, loffler2020vertebral, liebl2021computed}  & Vertebrae                                      & 28       & 64             & 16             & 80            \\
VerSe20\cite{sekuboyina2021verse, loffler2020vertebral, liebl2021computed}  & Vertebrae                                      & 28       & 48             & 13             & 61            \\
WORD\cite{luo2022word}                                                      & Thoracic and abdominal organs                  & 16       & 80             & 20             & 100           \\ \midrule
\textbf{Sum}                                                                                 & -                                              & -        & \textbf{4610} & \textbf{1162} & \textbf{5772} \\ \bottomrule
\end{tabular}
}
\end{table*}
% \FloatBarrier

\subsection{Datasets}
In addition to the detailed introduction of our datasets in Section 3 of the manuscript, we provide additional details of the M3D-Seg dataset in Table \ref{stab2}. M3D-Seg comprises 5,772 3D CTs and their corresponding masks collected from 25 public segmentation datasets. These datasets offer training and evaluation data for vision language positioning and segmentation tasks.

\section{Prompts and Templates}
\label{sec2}

In our work, we frequently utilize various prompts and templates to guide LLM in different tasks, including data generation, data checking, model evaluation, and task instruction. Figures \ref{sfig2} and \ref{sfig3} depict prompts for data generation, while Figure \ref{sfig4} illustrates prompts used for assessing the quality of generated VQA data. The evaluation of model performance using LLMs is demonstrated in Figure \ref{sfig5}. Additionally, instruction templates for different tasks are provided in Figures \ref{sfig6}, \ref{sfig7}, \ref{sfig8}, and \ref{sfig9}. Figure \ref{sfig10} presents samples from our term dictionary.

\begin{figure*}
\resizebox{1\textwidth}{!}{
\begin{tikzpicture}
    \node (example-textwidth-3) [draw, rounded corners,
                                 text width=1.3\linewidth,    % <---
                                 inner sep=12 pt]%
    { 
    \justify
    \small
You are a medical AI visual assistant that can analyze a single CT image. You receive the file name of the CT image and the medical diagnosis report. The report describes multiple abnormal lesions in the image.

\hfill

The task is to use the provided CT image and report information to create plausible 9 questions about the image.
Each question corresponds to four options, and these questions come from the following 5 aspects:

1). Planes (axial, sagittal, coronal);

2). CT phase (non-contrast, contrast, arterial phase, portal venous phase, venous phase, delayed phase, parenchymal phase, renal cortical phase, dual phase, renal excretory phase, mixed arteriovenous, myelography, etc.) or window ( bone, lung, window, etc.);

3). Organ;

4). Abnormality type or description;

5). Abnormality position;

\hfill

\textbf{Image:} $\{image\_file\_name\}$ \# It provides basic information about planes and phase.

\textbf{Report:} $\{text\}$ \# It provides detailed image findings and impressions.

\hfill

\textbf{Desired format:}\par

1). Planes

Question-1: ...? Choice: A. ... B. ... C. ... D. ... Answer: A. ...

2). CT phase

Question-2: ...? Choice: A. ... B. ... C. ... D. ... Answer: A. ...

3). Organ

Question-3: ...? Choice: A. ... B. ... C. ... D. ... Answer: A. ...

4). Abnormality type or description

Question-4: ...? Choice: A. ... B. ... C. ... D. ... Answer: A. ...

Question-5: ...? Choice: A. ... B. ... C. ... D. ... Answer: A. ...

Question-6: ...? Choice: A. ... B. ... C. ... D. ... Answer: A. ...

5). Abnormality position

Question-7: ...? Choice: A. ... B. ... C. ... D. ... Answer: A. ...

Question-8: ...? Choice: A. ... B. ... C. ... D. ... Answer: A. ...

Question-9: ...? Choice: A. ... B. ... C. ... D. ... Answer: A. ...

\hfill

Make the correct answers randomly distributed among the four choices.

If there is a true or false question, please ensure that the proportion of yes and no is equivalent. For example, Is ... ? Are ... ?, Do ... ?, Does ... ?, Did ... ?, Can ... ?.

Please do NOT ask directly what organs or abnormalities are visible in the image, as the answers are not unique. It would be best to use specific descriptions in your questions to ensure that other people can get an accurate answer even without providing choices.

Please be careful not to mention the file name and report. Always ask questions and answer as if directly looking at the image.
\hfill
};
\end{tikzpicture}
}
\caption{The prompt of VQA data generation. Specifically, we insert the image file name and report text into the placeholders (\{\}) within the prompt and feed it to LLM. Subsequently, we post-process the output of LLM to extract VQA data. Additionally, we observed that Qwen-72B\cite{qwen} and ChatGPT\cite{chatgpt} perform similarly in our data generation experiments, leading us to adopt the more cost-effective Qwen-72B model.}
\label{sfig2}
\end{figure*}

\begin{figure*}
\centering
\resizebox{1\textwidth}{!}{
\begin{tikzpicture}
    \node (example-textwidth-3) [draw, rounded corners,
                                 text width=1.4\linewidth,    % <---
                                 inner sep=12 pt]%
    { 
    \justify
    \normalsize

You are a medical AI visual assistant that can analyze a single CT image. Unfortunately you can't see the image but you can receive a diagnostic report of a local area in the CT image. The report describes the abnormal lesion in the image.

\hfill

The task is to use the provided report information to create plausible 6 questions and answers about the image for reasoning segmentation tasks

\hfill

\textbf{Report:} $\{text\}$ \# It provides detailed image findings and impressions.

\hfill

Questions and answers need to be structured from the report. But don't mention the report in Q\&A. The question needs to be about a specific lesion area and requires segmentation of this area. The answer needs to use only one $<$SEG$>$ symbol to refer to the segmentation area and provide a text explanation. 

\hfill

There are two types of questions: one type of question is answered and segmented based on description information, and the other type of question requires reasoning based on general and medical knowledge to obtain answers and segmentation.

\hfill

\textbf{Example:}

1). Description-based

Question-1: Please segment where the liver cyst appears in the image. Answer: Sure, it is [SEG] on the upper right side of the liver.

2). Reasoning-based

Question-1: Can you segment the unusual part in this image and explain why? Answer: Sure, it is [SEG]. In the image, the unusual part is ...

Question-2: What can make the woman stand higher? Please output segmentation mask and explain why. Answer: Sure, [SEG]. The woman is standing higher by using ...

Question-3: If there are any lesions in the largest human body organ in the image, please segment them. Answer: The largest organ is the liver, where liver tumors are present, and the region is the $<$SEG$>$.

\hfill

\textbf{Desired output format:}

1). Description-based

Question-1: ...? Answer: ...

Question-2: ...? Answer: ...

Question-3: ...? Answer: ...

2). Reasoning-based

Question-4: ...? Answer: ...

Question-5: ...? Answer: ...

Question-6: ...? Answer: ...

\hfill

Please construct a total of 6 sets of question and answer pairs according to the desired format, 3 sets of each type.

Using specific descriptions in your questions would ensure others can get an accurate answer.

Always ask questions and answer as if directly looking at the image.

};
\end{tikzpicture}
}
\caption{The prompt of data generation for referring expression segmentation. Specifically, we insert the report description for a mask into the placeholders (\{\}) within the prompt and feed it to LLM. Subsequently, we post-process the output of LLM to extract instructions. This involves generating diverse descriptions and inferential questions from simple diagnostic reports. Qwen-72B is selected for this task due to its efficiency and performance.
}
\label{sfig3}
\end{figure*}

\begin{figure*}
\centering
\resizebox{0.8\textwidth}{!}{
\begin{tikzpicture}
    \node (example-textwidth-3) [draw, rounded corners,
                                 text width=1.2\linewidth-24pt,    % <---
                                 inner sep=12 pt]%
    { 
    \justify
    \normalsize

You are a medical AI assistant. Please provide answers and help based on the following questions.

\hfill

This is a question from the visual question-answering dataset. Questions are generated based on information from images and reports, and the generated data inevitably contains certain errors. 

\hfill

Please use the following information to determine whether the content described in the question is consistent with the text report and whether the answer is correct.

\hfill

\textbf{Image Path:} $\{img\_file\_name\}$ \# It provides basic information about planes and phase.

\textbf{Report:} $\{text\}$ \# It provides detailed image findings and impressions.

\textbf{Question:} $\{question\}$

\textbf{Choices:} A. $\{choice\_A\}$ B. $\{choice\_B\}$ C. $\{choice\_C\}$ D. $\{choice\_D\}$

\textbf{Answer Choice:} $\{answer\_choice\}$. $\{answer\}$

\hfill

If there is an error, please answer 'NO' first and give a more reasonable question and answer. If it is basically correct, answer 'Yes' directly. Do not give redundant answers.

};
\end{tikzpicture}
}
\caption{The prompt for VQA data checking. Manual verification of large datasets is highly resource-intensive, especially in medical scenarios. Therefore, we employ LLM to automatically check the generated data. Specifically, we insert the generated data (questions, choices, and answers) and original data (image file names, report text) into the placeholders (\{\}) within the prompt and feed them to LLM. Subsequently, we assess the validity of the data by analyzing LLM's output. Qwen-72B is selected for this task due to its efficiency and performance.}
\label{sfig4}
\end{figure*}

\begin{figure*}
\centering
\resizebox{0.8\textwidth}{!}{
\begin{tikzpicture}
    \node (example-textwidth-3) [draw, rounded corners,
                                 text width=\linewidth-24pt,    % <---
                                 inner sep=12 pt]%
    { 
    \justify
    \small
You are an AI assistant, please evaluate based on the following.

\hfill

Please refer to the ground truth and prediction based on the following two paragraphs, identify the aspects mentioned in the ground truth, and calculate the percentage of these aspects that are either correctly mentioned or partially matched in the prediction, scoring from 0 to 100.

\hfill

\textbf{Ground truth:} $\{answer\}$ \# Reference text

\textbf{Prediction:} $\{prediction\}$ \# Generated text

\hfill

Please follow the output format:

Score: xx. The reason is ......

};
\end{tikzpicture}
}
\caption{The prompt for model evaluation. Traditional metrics struggle to evaluate the model's performance semantically. Therefore, we introduce an LLM-based metric for evaluation. In detail, we insert ground truth and predictions into the placeholders (\{\}) within the prompt and input them to LLM. Subsequently, we extract the absolute scores directly from the output of the LLM. Qwen-72B is chosen for this task due to its efficiency and performance.}
\label{sfig5}
\end{figure*}

\begin{figure*}
\centering
\begin{tikzpicture}
    \node (example-textwidth-3) [draw, rounded corners,
                                 text width=0.9\linewidth,    % <---
                                 inner sep=12 pt]%
    { 
    \justify
    \small
\textbf{Report Generation:} \par
\begin{itemize}
\item[\textbullet] Can you provide a caption consists of findings for this medical image?

\item[\textbullet] Describe the findings of the medical image you see.

\item[\textbullet] Please caption this medical scan with findings.

\item[\textbullet] What is the findings of this image?

\item[\textbullet] Describe this medical scan with findings.

\item[\textbullet] Please write a caption consists of findings for this image.

\item[\textbullet] Can you summarize with findings the images presented?

\item[\textbullet] Please caption this scan with findings.

\item[\textbullet] Please provide a caption consists of findings for this medical image.

\item[\textbullet] Can you provide a summary consists of findings of this radiograph?

\item[\textbullet] What are the findings presented in this medical scan?

\item[\textbullet] Please write a caption consists of findings for this scan.

\item[\textbullet] Can you provide a description consists of findings of this medical scan?

\item[\textbullet] Please caption this medical scan with findings.

\item[\textbullet] Can you provide a caption consists of findings for this medical scan?
\end{itemize}
};
\end{tikzpicture}
\caption{Examples of instructions for report generation. These instructions typically include prompts or guidelines for generating specific sections or content within the medical reports. These instructions, along with corresponding images, are input into the MLLM together to facilitate the report generation process.
}
\label{sfig6}
\end{figure*}

\begin{figure*}
\centering
\resizebox{0.95\textwidth}{!}{
\begin{tikzpicture}
    \node (example-textwidth-3) [draw, rounded corners,
                                 text width=1.3\linewidth,    % <---
                                 inner sep=12 pt]%
    { 
    \justify
    \small
\textbf{Referring Expression Comprehension:}

\hfill

Category Questions:

\begin{itemize}
\item[\textbullet] Can you find the \{\} in this image? Give coordinates.

\item[\textbullet]Can you find \{\} in this image? Please output the coordinates.

\item[\textbullet]Please bounding the \{\} by box in this image.

\item[\textbullet]Where is \{\} in this image? Please respond with a bounding box.

\item[\textbullet]Where is \{\} in this image? Please output the box.

\item[\textbullet]Can you locate the \{\} in this image? Please output its coordinates.

\item[\textbullet]Could you mark the \{\} by bounding box in this image?

\item[\textbullet]Where can I find the \{\} in this image? Please provide its bounding box.

\item[\textbullet]Identify the indicated \{\} in this image. Please provide the coordinates of its bounding box.
\end{itemize}

Answers:
\begin{itemize}
\item[\textbullet] Coordinates are \{\}.

\item[\textbullet]Sure, \{\}.

\item[\textbullet]Sure, it is \{\}.

\item[\textbullet]Sure, the bounding box is \{\}.

\item[\textbullet]\{\}.

\item[\textbullet]Here are the coordinates: \{\}.

\item[\textbullet]Of course, it's located at \{\}.

\item[\textbullet]The bounding box is given by \{\}.

\item[\textbullet]The box is \{\}.
\end{itemize}

Description Questions:

\begin{itemize}
\item[\textbullet]Description: \{\} Please answer and find it by box based on the above description.

\item[\textbullet]Definition: \{\} Please answer and show the bounding box based on the above definition.

\item[\textbullet]Description: \{\} Can you answer and find it by coordinates based on the description?

\item[\textbullet]Definition: \{\} Please output the bounding box and answer based on the definition.

\item[\textbullet]Description: \{\} Respond and locate it using a bounding box according to the description.

\item[\textbullet]Definition: \{\} Please provide an answer and display the bounding box according to the given definition.

\item[\textbullet]Description: \{\} Can you identify and locate it by coordinates, following the provided description or definition?

\item[\textbullet]Definition: \{\} Please output the bounding box and provide an answer based on the provided definition.

\item[\textbullet]Based on the description or definition, please respond to \{\} and indicate its location with a bounding box.

\end{itemize}

Answers:

\begin{itemize}
\item[\textbullet] The target is \{\} and the coordinates is \{\}.

\item[\textbullet]The category is \{\} and the bounding box is \{\}.

\item[\textbullet]It is \{\}, \{\}.

\item[\textbullet]\{\}, \{\}

\item[\textbullet]The target is identified as \{\} and its coordinates are \{\}.

\item[\textbullet]The category is \{\}, the bounding box is provided as \{\}.

\item[\textbullet]It is characterized by \{\}, with coordinates \{\}.

\item[\textbullet]The identified attributes are \{\}, \{\}.

\item[\textbullet]Describing it as \{\}, the corresponding box is \{\}.

\end{itemize}

};
\end{tikzpicture}
}
\caption{Instruction templates for referring expression comprehension. These templates guide the construction of instruction data for the referring expression comprehension task. The data for this task is sourced from M3D-Seg, a segmentation dataset providing categories and bounding boxes. In category questions, categories are inserted into question templates' placeholders (\{\}) as input, while bounding boxes are inserted into answer templates' placeholders (\{\}) as output. In description questions, categories are converted into descriptions using the term dictionary. These instruction templates facilitate the generation of instruction data for referring expression comprehension.
}
\label{sfig7}
\end{figure*}

\begin{figure*}
\centering
\resizebox{\textwidth}{!}{
\begin{tikzpicture}
    \node (example-textwidth-3) [draw, rounded corners,
                                 text width=1.3\linewidth,    % <---
                                 inner sep=12 pt]%
    { 
    \justify
    \small
\textbf{Referring Expression Generation:}

\hfill

Category Questions:

\begin{itemize}
\item[\textbullet] What target is present within the coordinates \{\} ?

\item[\textbullet]Does the bounding box \{\} contain any target?

\item[\textbullet]Within the specified region \{\}, what target is present?

\item[\textbullet]Do you know what it is in the bounding box \{\}?

\item[\textbullet]What is it in this region \{\}?

\item[\textbullet]What object is located within the coordinates \{\}?

\item[\textbullet]Within the specified area \{\}, what object can be found?

\item[\textbullet]Can you identify the object within the bounding box \{\}?

\item[\textbullet]What object is present in this region \{\}?
\end{itemize}

Answer:
\begin{itemize}
\item[\textbullet] The target is \{\}.

\item[\textbullet]Sure, the bounding box contains \{\}.

\item[\textbullet]Sure, it is \{\}.

\item[\textbullet]Sure, \{\} is in the bounding box.

\item[\textbullet]\{\}.

\item[\textbullet]The object is \{\}.

\item[\textbullet]Of course, it's \{\}.

\item[\textbullet]Certainly, \{\} can be found in the bounding box.

\item[\textbullet]Yes, the bounding box includes \{\}.
\end{itemize}

Description Questions:

\begin{itemize}
\item[\textbullet]Please describe the target and its function based on the box \{\} in the image.

\item[\textbullet]Do you know what is it in this bounding box \{\}? Answer and explain it.

\item[\textbullet]What's the target in the bounding box \{\}? What function does it have?

\item[\textbullet]What is the area marked with a box \{\} in the image? Can you explain it?

\item[\textbullet]Could you describe the object and its purpose within the bounding box \{\} in the image?

\item[\textbullet]Can you identify and describe the object within this bounding box \{\}? Please explain.

\item[\textbullet]What is the object located in the bounding box \{\}? Could you explain its function?

\item[\textbullet]Could you describe the area outlined by the box \{\} in the image? Please explain its significance.

\end{itemize}

Answer:

\begin{itemize}
\item[\textbullet] Sure, it is \{\}. \{\}.

\item[\textbullet]The category is \{\}. \{\}.

\item[\textbullet]It is \{\}, \{\}.

\item[\textbullet]\{\}, \{\}

\item[\textbullet]The target is identified as \{\} and its description is \{\}.

\item[\textbullet]The category is \{\}. Description: \{\}.

\item[\textbullet]It is characterized by \{\}, \{\}.

\item[\textbullet]The identified attributes are \{\}, \{\}.

\item[\textbullet]Sure, it is \{\}. Describing it as \{\}.

\end{itemize}

};
\end{tikzpicture}
}
\caption{Instruction templates for referring expression generation. These templates facilitate the construction of instruction data for the referring expression generation task. In category questions, bounding boxes are inserted into question templates' placeholders (\{\}) as input, while categories are inserted into answer templates' placeholders (\{\}) as output. Similarly, in description questions, categories are converted into descriptions using the term dictionary. The model is expected to output both the target and its description as answers.
}
\label{sfig8}
\end{figure*}

\begin{figure*}
\resizebox{0.95\textwidth}{!}{
\begin{tikzpicture}
    \node (example-textwidth-3) [draw, rounded corners,
                                 text width=1.25\linewidth,    % <---
                                 inner sep=12 pt]%
    { 
    \justify
    \small
    
\textbf{Semantic Segmentation:}

\hfill

Question:
\begin{itemize}
\item[\textbullet]Can you segment the \{\} in this image?

\item[\textbullet]Can you segment \{\} in this image? Please output the mask.
\item[\textbullet]Please segment the \{\} in this image.

\item[\textbullet]What is \{\} in this image? Please respond with segmentation mask.

\item[\textbullet]What is \{\} in this image? Please output segmentation mask.
\item[\textbullet]Could you provide a segmentation for the \{\}?

\item[\textbullet]Segment \{\} from this image and provide the mask, please.

\item[\textbullet]Please provide a segmentation mask for the \{\} in this image.

\item[\textbullet]Can you identify and segment the \{\} in this image?
\end{itemize}

Answer:

\begin{itemize}
\item[\textbullet]It is [SEG].

\item[\textbullet]Sure, [SEG].

\item[\textbullet]Sure, it is [SEG].

\item[\textbullet]Sure, the segmentation result is [SEG].

\item[\textbullet]The segmentation indicates [SEG].

\item[\textbullet]According to the segmentation, it is [SEG].

\item[\textbullet]The segmentation reveals [SEG].

\item[\textbullet]The segmentation suggests [SEG].

\item[\textbullet]From the segmentation, it appears to be [SEG].
\end{itemize}

\hfill

\noindent \textbf{Referring Expression Segmentation:}

\hfill

Question:

\begin{itemize}
\item[\textbullet] Description: \{\} Please answer and segment based on the above description.

\item[\textbullet] Definition: \{\} Please answer and segment based on the above definition.

\item[\textbullet] Description: \{\} Can you answer and segment it based on the above description or definition.

\item[\textbullet]Definition: \{\} Please output segmentation mask and answer based on the above description or definition.

\item[\textbullet]Provided description: \{\} Please segment accordingly.

\item[\textbullet]Given definition: \{\} Please provide segmentation and answer according to it.

\item[\textbullet]The description provided is: \{\} Now, segment it and provide your answer.

\item[\textbullet]Based on the provided definition: \{\} Please segment and provide your response.

\item[\textbullet]Describing the object as: \{\} Can you segment it accordingly?
\end{itemize}

Answer:
\begin{itemize}
\item[\textbullet] The target is \{\} and the segmentation mask is [SEG].

\item[\textbullet]The category is \{\} and the mask is [SEG].

\item[\textbullet]It is \{\}, [SEG].

\item[\textbullet]Identified as \{\}, here is the segmentation: [SEG].

\item[\textbullet]Categorized as \{\}, the segmentation is: [SEG].

\item[\textbullet]The class is \{\}, and the corresponding segmentation is: [SEG].

\item[\textbullet]Regarding the classification, it is \{\}, and the segmentation is: [SEG].

\item[\textbullet]Classified as \{\}, here's the segmentation: [SEG].

\end{itemize}

};
\end{tikzpicture}
}
\caption{Instruction templates for segmentation tasks. In semantic segmentation, categories are inserted into question templates' placeholders (\{\}) as input. For referring expression segmentation, descriptions are inserted into question templates' placeholders (\{\}) as input. In both cases, all answers include a special token [SEG], which instructs the segmentation module. This token is crucial for guiding the segmentation process based on the provided input.
}
\label{sfig9}
\end{figure*}

\begin{figure*}
\resizebox{\textwidth}{!}{
\begin{tikzpicture}
    \node (example-textwidth-3) [draw, rounded corners,
                                 text width=1.2\linewidth,    % <---
                                 inner sep=12 pt]%
    { 
    \justify
    \small
    
\{

"liver": [

    \hspace{20pt} "Primary organ responsible for detoxifying the blood by removing harmful substances.",
    
    \hspace{20pt} "Produces bile, a fluid that aids in the digestion and absorption of fats.",
    
    \hspace{20pt} "Stores and regulates glycogen, a crucial energy reserve for the body.",
    
    \hspace{20pt} "Synthesizes proteins necessary for blood clotting and immune system function.",
    
    \hspace{20pt} "Plays a central role in metabolism, including the breakdown of carbohydrates and fats.",
    
    \hspace{20pt} "Large organ in the upper right abdomen with various metabolic functions.",
    
    \hspace{20pt} ......],
    
"left lung": [

    \hspace{20pt} "Organ located on the left side of the chest involved in respiration.",
    
    \hspace{20pt} "Respiratory organ situated in the left thoracic cavity.",
    
    \hspace{20pt} "Lung found on the left side of the body responsible for breathing.",
    
    \hspace{20pt} "Pulmonary structure on the left side of the chest responsible for gas exchange.",
    
    \hspace{20pt} "Left-sided respiratory organ essential for oxygen exchange.",
    
    \hspace{20pt} "Organ situated in the left thorax responsible for oxygenating blood.",
    
    \hspace{20pt} "Lung located in the left hemithorax involved in ventilation.",
    
    \hspace{20pt} ......],
    
"kidney": [

    \hspace{20pt} "Pair of organs responsible for filtering waste from the blood.",
    
    \hspace{20pt} "Organ duo involved in removing waste and excess fluids from the body.",
    
    \hspace{20pt} "Pair of bean-shaped organs essential for regulating bodily fluids.",
    
    \hspace{20pt} "Organs crucial for filtering blood and producing urine.",
    
    \hspace{20pt} "Pair of vital organs filtering blood and maintaining fluid balance.",
    
    \hspace{20pt} "Bean-shaped organs integral to waste removal and urine production.",
    
    \hspace{20pt} "Organs vital for removing toxins and excess fluids from the body.",
    
    \hspace{20pt} ......],
     
"heart": [

    \hspace{20pt} "Organ responsible for pumping blood throughout the body.",
    
    \hspace{20pt} "Muscular organ that circulates blood throughout the circulatory system.",
    
    \hspace{20pt} "Vital organ that pumps oxygenated blood to tissues and organs.",
    
    \hspace{20pt} "Primary pump of the circulatory system, supplying oxygen to tissues.",
    
    \hspace{20pt} "Central organ of the cardiovascular system, propelling blood throughout the body.",
    
    \hspace{20pt} "Main organ of the circulatory system, distributing nutrients and oxygen.",
    
    \hspace{20pt} ......],
    
"liver tumor": [

   \hspace{20pt} "Abnormal growth in liver tissue.",
    
    \hspace{20pt} "Mass of cells forming in the liver.",
    
    \hspace{20pt} "Neoplastic lesion found in the liver.",
    
    \hspace{20pt} "Pathological growth occurring in liver tissue.",
    
    \hspace{20pt} "Uncontrolled cell proliferation in the liver.",
    
    \hspace{20pt} "Anomaly of tissue growth within the liver.",
    
    \hspace{20pt} ......],
    
......

\noindent\}
};
\end{tikzpicture}
}
\caption{Examples from the term dictionary. The term dictionary contains multiple descriptions for each medical term. These descriptions are generated through ChatGPT. With numerous medical terms included, this dictionary is crucial in transforming semantic categories into detailed descriptions. These descriptions are essential for facilitating vision language positioning and segmentation tasks.
}
\label{sfig10}
\end{figure*}

\section{Qualitative Analysis}
\label{sec3}
\begin{figure*}[htbp]
\centering
\includegraphics[width=0.9\textwidth]{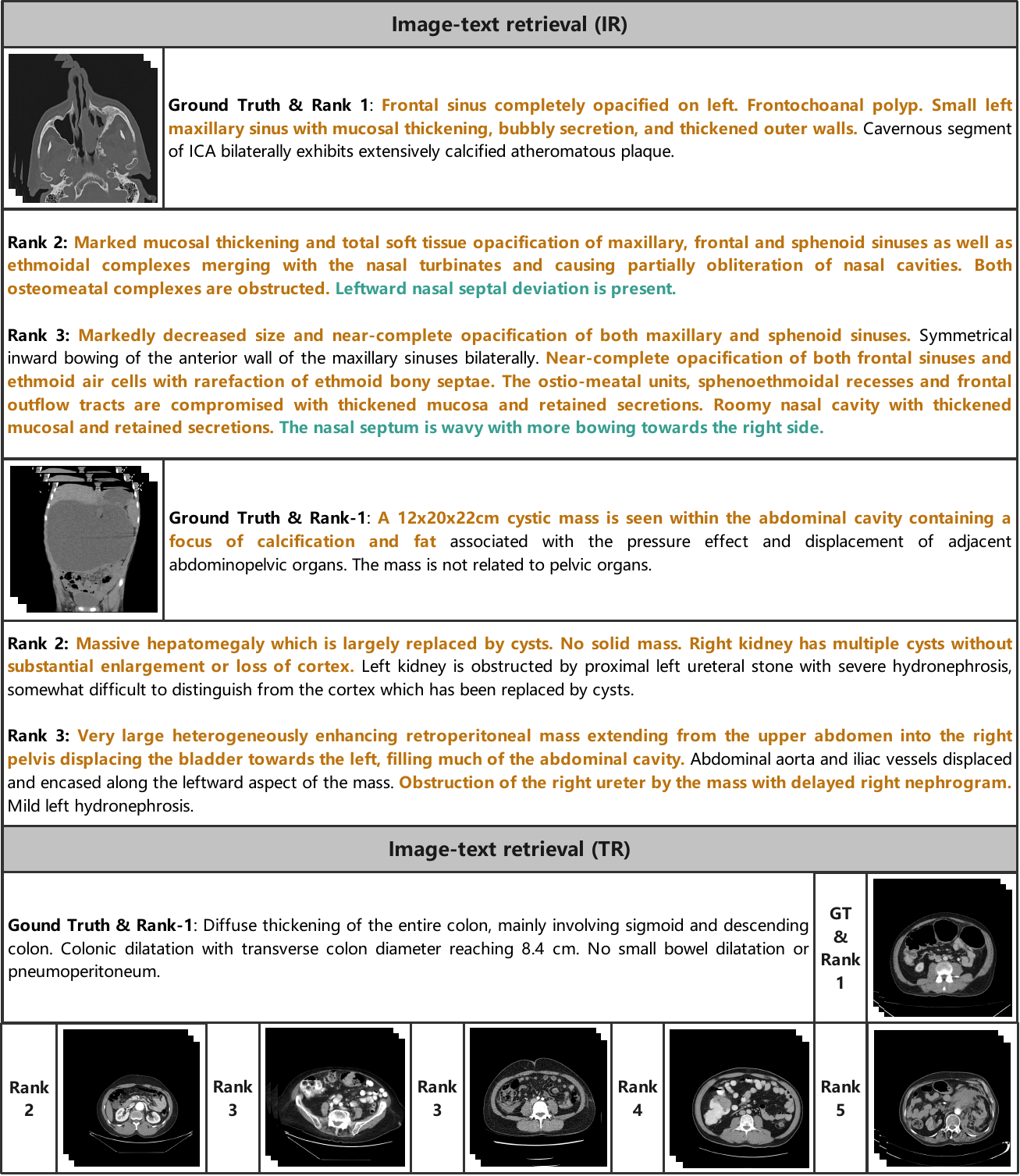}
\caption{Qualitative comparisons on image-text retrieval. In each case, text with the same color represents identical content, while text with different colors signifies different content. The top-ranked samples exhibit similarities to ground truth (GT) in both image-to-text retrieval (IR) and text-to-image retrieval (TR) tasks.} 
\label{sfig11}
\end{figure*}

\begin{figure*}[htbp]
\centering
\includegraphics[width=0.95\textwidth]{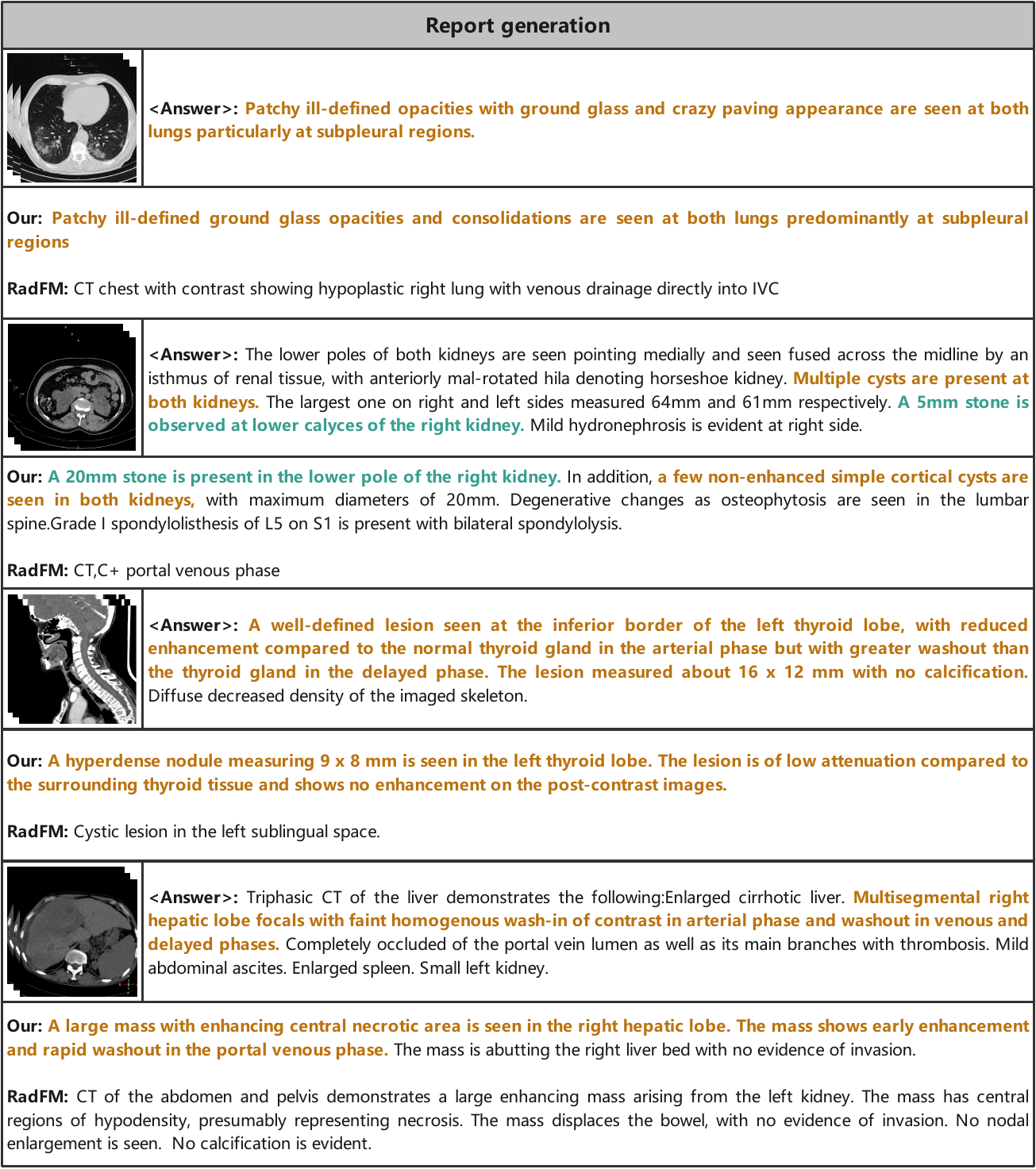}
\caption{Qualitative comparisons with another method in report generation. Text with the same color indicates identical content, while different colors signify differing content. Our model exhibits superior performance to RadFM by generating more answer-identical content. We attempted to include GPT-4V in the comparison but encountered limitations, as it struggled to generate medical-related diagnostic recommendations.} 
\label{sfig12}
\end{figure*}

\begin{figure*}[htbp]
\centering
\includegraphics[width=0.95\textwidth]{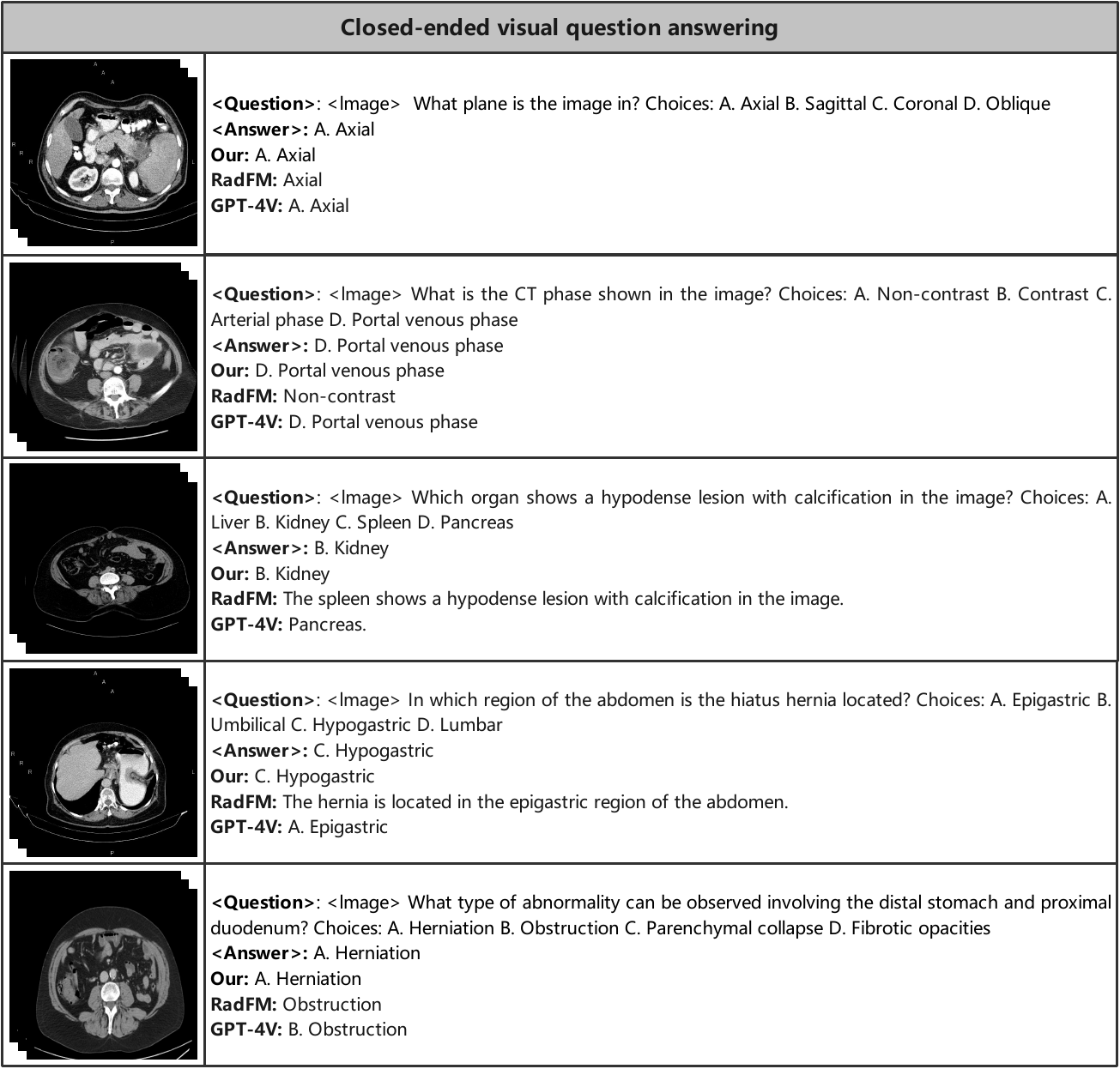}
\caption{Qualitative comparisons with other methods in closed-ended VQA. We compare methods across five types: plane, phase, organ, abnormality, and location, highlighting the superiority of our approach. In closed-ended VQA, GPT-4V sometimes relaxes the restrictions, because we provide choices that may turn medical advice into a multiple-choice question.} 
\label{sfig13}
\end{figure*}

\begin{figure*}[htbp]
\centering
\includegraphics[width=0.95\textwidth]{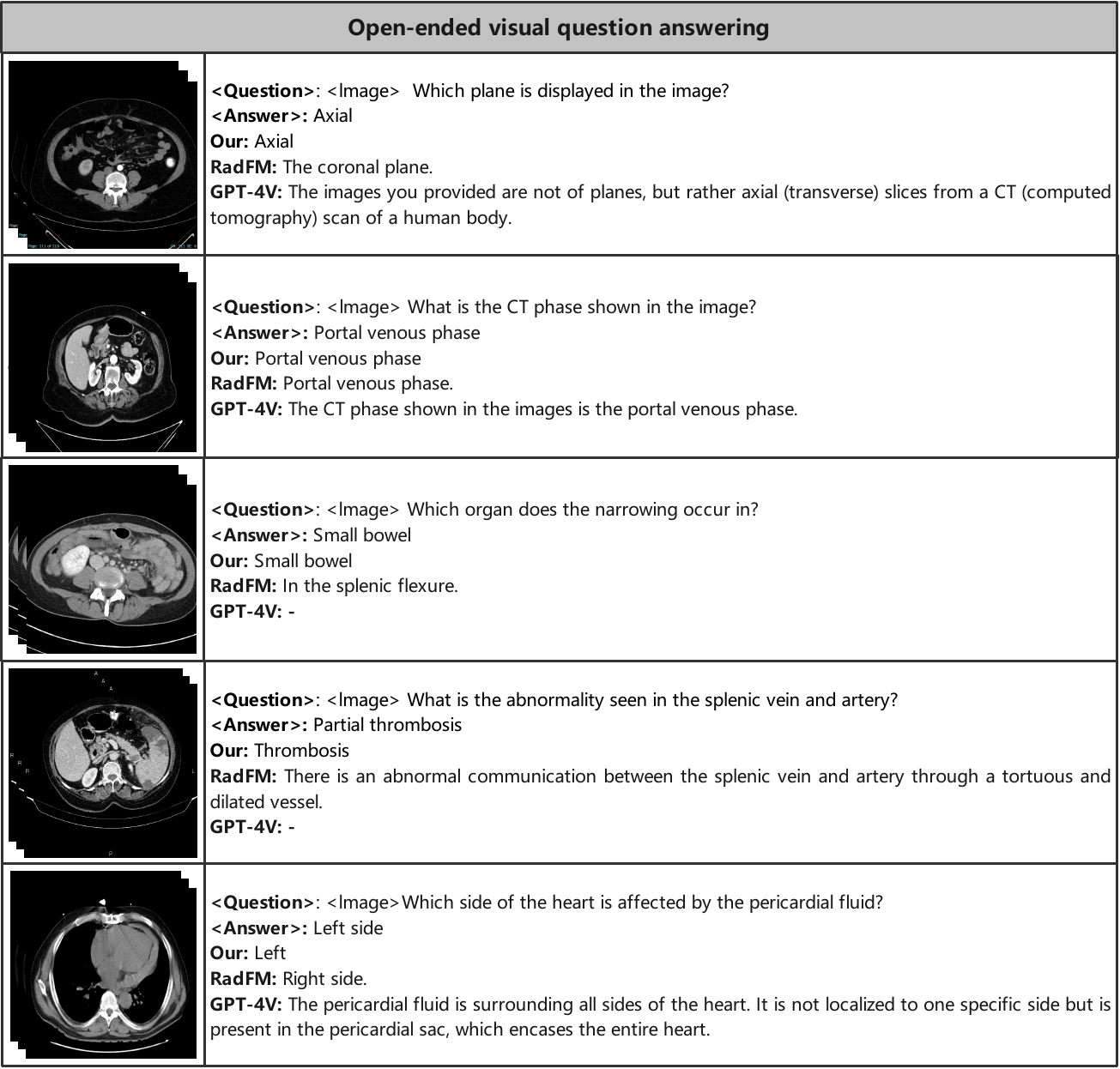}
\caption{Qualitative comparisons with other methods in open-ended VQA. Similarly, our method demonstrates superior performance across five types. However, questions related to abnormality topics in open-ended VQA remain restricted by GPT-4V. In cases where no valid answer can be obtained, "-" is used to indicate this limitation.} 
\label{sfig14}
\end{figure*}

\begin{figure*}[htbp]
\centering
\includegraphics[width=1\textwidth]{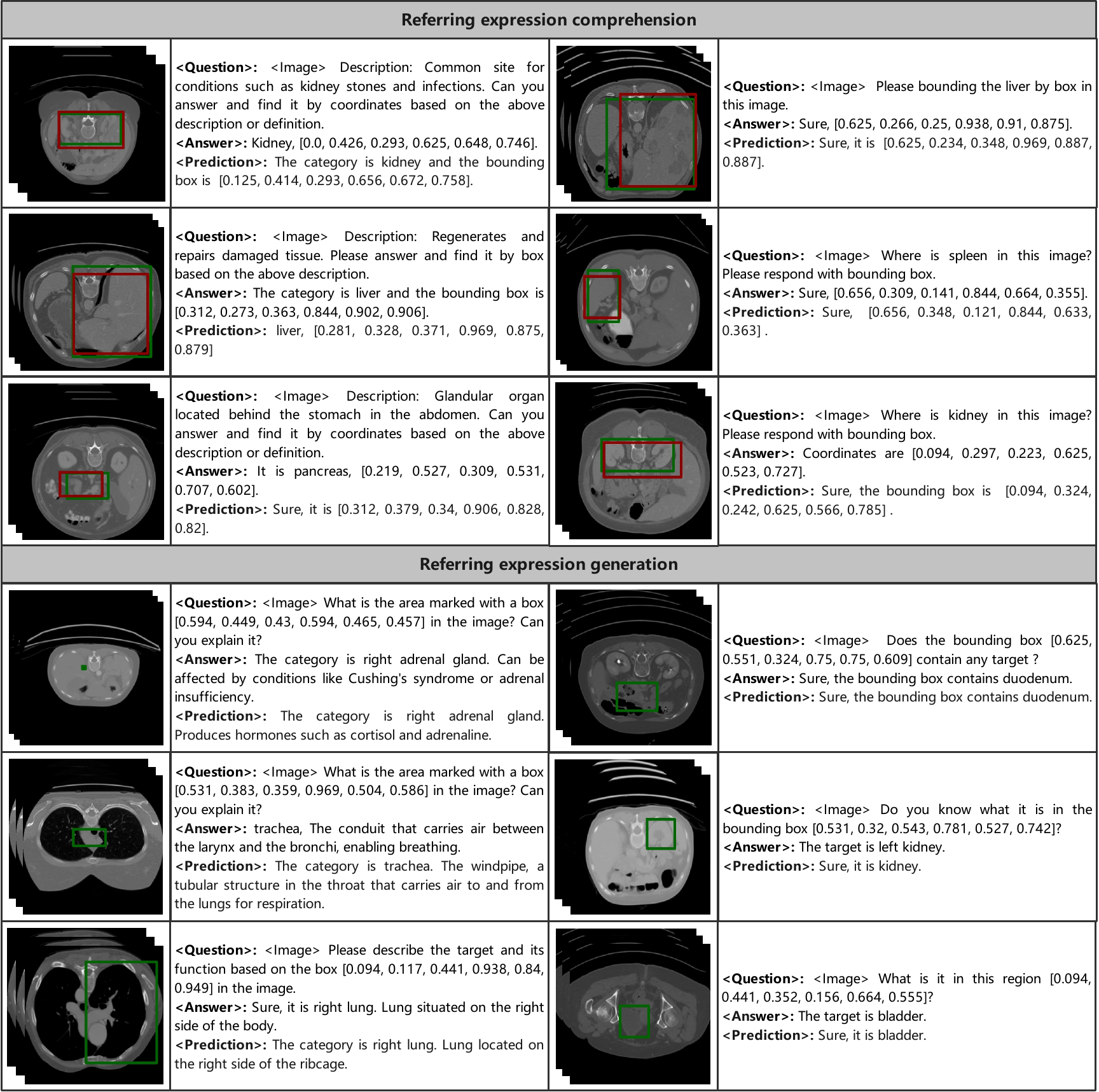}
\caption{Qualitative analysis on positioning tasks. We demonstrate two task forms: box output and box input, representing referring expression comprehension and referring expression generation, respectively. This demonstrates our model's effectiveness in completing the vision language positioning task. In the visualizations, the green box represents the ground truth, while the red box represents the prediction.} 
\label{sfig15}
\end{figure*}

\begin{figure*}[htbp]
\centering
\includegraphics[width=1\textwidth]{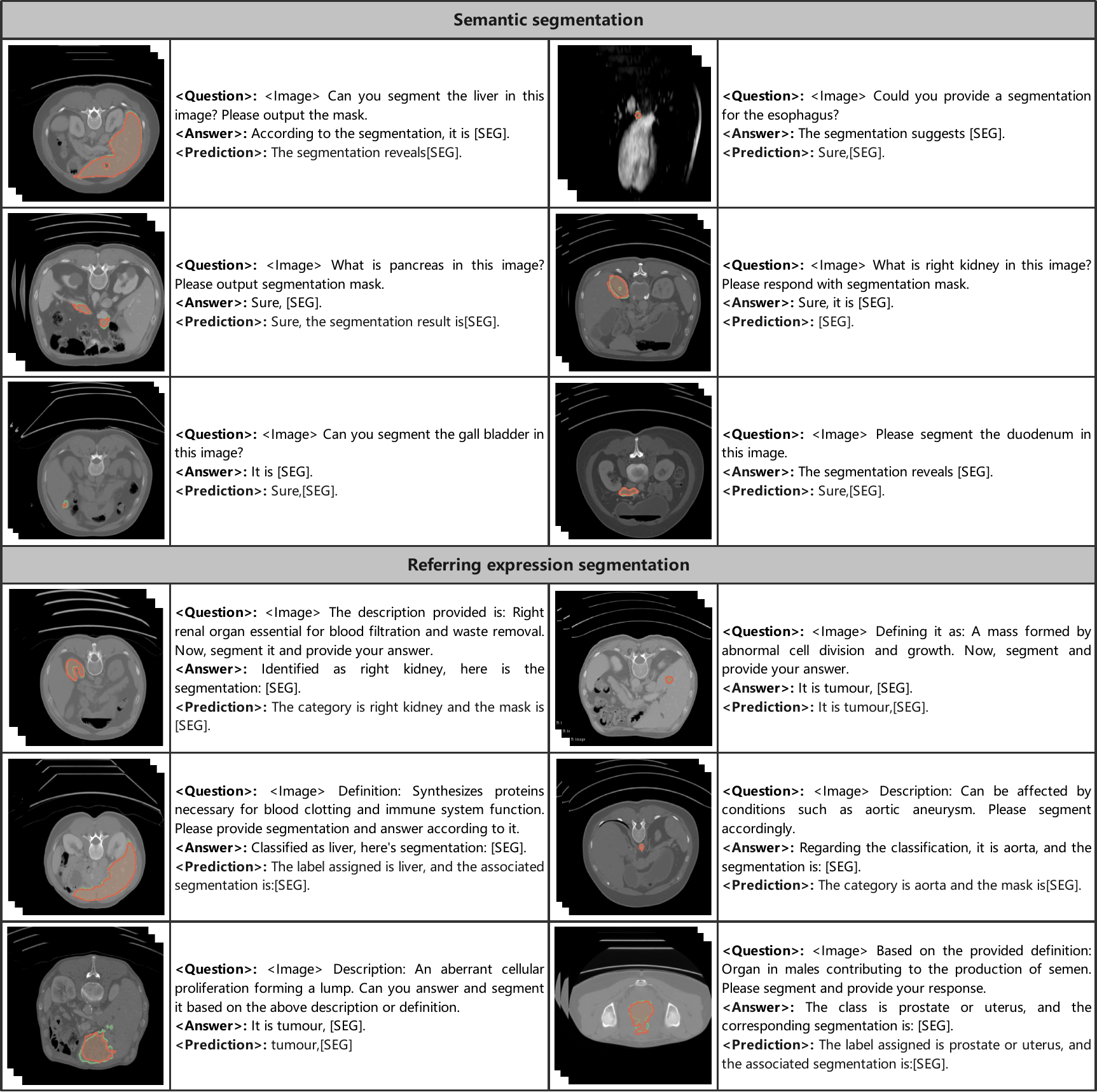}
\caption{Qualitative analysis on segmentation tasks. We show two task forms: semantic segmentation and referring expression segmentation, highlighting our model's proficiency in segmentation tasks. In the visualizations, the green mask represents the ground truth, while the red mask represents the prediction.} 
\label{sfig16}
\end{figure*}

To further demonstrate our model's performance and generalist ability on 3D multi-modal medical tasks, we add additional qualitative analysis on 8 tasks: image-text retrieval (Figure \ref{sfig11}), report generation (Figure \ref{sfig12}), closed-ended VQA (Figure \ref{sfig13}), open-ended VQA (Figure \ref{sfig14}), referring expression comprehension (Figure \ref{sfig15}), referring expression generation (Figure \ref{sfig15}), semantic segmentation, (Figure \ref{sfig16}) and referring expression segmentation (Figure \ref{sfig16}). 

\end{document}